\documentclass[10pt,twocolumn,letterpaper]{article}

\usepackage[pagenumbers]{cvpr} % To force page numbers, e.g. for an arXiv version

\renewcommand{\paragraph}[1]{\vspace{.5em}\noindent\textbf{#1.}}

\usepackage{soul}
\setuldepth{foobar}

\usepackage{adjustbox}
\usepackage{multirow}
\usepackage{makecell}
\usepackage{tcolorbox}
\usepackage{listings}
\usepackage{graphicx}
\usepackage{caption}
\usepackage{float}
\usepackage{dblfloatfix} 

\usepackage[table]{xcolor} % for \cellcolor
\usepackage{booktabs}      % for nicer tables
\usepackage{pgf}           % for calculations
\usepackage{fontawesome5}
\usepackage{pifont}
\usepackage{svg} %moritz
\usepackage{longtable}
\usepackage{booktabs}
\usepackage{multirow}
\usepackage{adjustbox}
\usepackage{upquote}
\usepackage{tikz}
\usepackage{pgf-pie}
\newcommand{\cmark}{\ding{51}}%
\newcommand{\xmark}{\ding{55}}%

\usepackage{textcomp}
\makeatletter
\renewcommand{\@fnsymbol}[1]{*}  % Change to bold and large asterisk
\renewcommand{\thefootnote}{\fnsymbol{footnote}}  % Use symbols instead of numbers
\makeatother

\makeatletter
\newif\ifcvpr@inappendix
\cvpr@inappendixfalse

\newcommand{\CVPRAppendixTOCMarker}{}

\newcommand{\MarkAppendixInTOC}{%
  \addtocontents{toc}{\protect\CVPRAppendixTOCMarker}%
}

\newcommand{\AppendixOnlyTOC}{%
  \begingroup
    \let\cvpr@orig@section\l@section
    \let\cvpr@orig@subsection\l@subsection
    \let\cvpr@orig@subsubsection\l@subsubsection
    \let\cvpr@orig@paragraph\l@paragraph
    \let\cvpr@orig@subparagraph\l@subparagraph
    \renewcommand{\l@section}[2]{\ifcvpr@inappendix \cvpr@orig@section{##1}{##2}\fi}
    \renewcommand{\l@subsection}[2]{\ifcvpr@inappendix \cvpr@orig@subsection{##1}{##2}\fi}
    \renewcommand{\l@subsubsection}[2]{\ifcvpr@inappendix \cvpr@orig@subsubsection{##1}{##2}\fi}
    \renewcommand{\l@paragraph}[2]{\ifcvpr@inappendix \cvpr@orig@paragraph{##1}{##2}\fi}
    \renewcommand{\l@subparagraph}[2]{\ifcvpr@inappendix \cvpr@orig@subparagraph{##1}{##2}\fi}
    \renewcommand{\CVPRAppendixTOCMarker}{\global\cvpr@inappendixtrue}
    \tableofcontents
  \endgroup
}
\makeatother

\definecolor{mri}{RGB}{66,133,244}    % Blue-ish for MRI
\definecolor{ct}{RGB}{220,53,69}      % Red-ish for CT
\definecolor{pet}{RGB}{255,140,0}     % Orange for PET

\newcommand{\MRI}{\textcolor{mri}{\textbf{\texttt{MRI}}}}
\newcommand{\CT}{\textcolor{ct}{\textbf{\texttt{CT}}}}
\newcommand{\PET}{\textcolor{pet}{\textbf{\texttt{PET}}}}

\definecolor{goodred}{HTML}{e67c73}
\definecolor{goodyellow}{HTML}{ffd666}
\definecolor{goodgreen}{HTML}{57bb8a}

\pgfmathsetmacro{\minval}{0}  % Minimum value in the table
\pgfmathsetmacro{\maxval}{73}  % Maximum value in the table
\pgfmathsetmacro{\midval}{(\minval+\maxval)/2}  % Midpoint value for the yellow transition

\newcommand{\colorrange}[1]{%
    \pgfmathsetmacro{\PercentColor}{100.0*(#1-\minval)/(\maxval-\minval)}%
    \ifdim #1 pt < \minval pt
        \cellcolor{goodred}{#1}%
    \else
        \ifdim #1 pt > \maxval pt
            \cellcolor{goodgreen}{#1}%
        \else
            \ifdim #1 pt < \midval pt
                \pgfmathsetmacro{\PercentLow}{200.0*(#1-\minval)/(\maxval-\minval)}%
                \xdef\PercentLow{\PercentLow}%
                \cellcolor{goodyellow!\PercentLow!goodred}{#1}%
            \else
                \pgfmathsetmacro{\PercentHigh}{200.0*(#1-\midval)/(\maxval-\minval)}%
                \xdef\PercentHigh{\PercentHigh}%
                \cellcolor{goodgreen!\PercentHigh!goodyellow}{#1}%
            \fi
        \fi
    \fi
}

\pgfmathsetmacro{\minvala}{0}  % Minimum value in the table
\pgfmathsetmacro{\maxvala}{91}  % Maximum value in the table
\pgfmathsetmacro{\midvala}{(\minvala+\maxvala)/2}  % Midpoint value for the yellow transition

\newcommand{\colorrangea}[1]{%
    \pgfmathsetmacro{\PercentColor}{100.0*(#1-\minvala)/(\maxvala-\minvala)}%
    \ifdim #1 pt < \minvala pt
        \cellcolor{goodred}{#1}%
    \else
        \ifdim #1 pt > \maxvala pt
            \cellcolor{goodgreen}{#1}%
        \else
            \ifdim #1 pt < \midvala pt
                \pgfmathsetmacro{\PercentLow}{200.0*(#1-\minvala)/(\maxvala-\minvala)}%
                \xdef\PercentLow{\PercentLow}%
                \cellcolor{goodyellow!\PercentLow!goodred}{#1}%
            \else
                \pgfmathsetmacro{\PercentHigh}{200.0*(#1-\midvala)/(\maxvala-\minvala)}%
                \xdef\PercentHigh{\PercentHigh}%
                \cellcolor{goodgreen!\PercentHigh!goodyellow}{#1}%
            \fi
        \fi
    \fi
}

\pgfmathsetmacro{\minvalb}{0}  % Minimum value in the table
\pgfmathsetmacro{\maxvalb}{65}  % Maximum value in the table
\pgfmathsetmacro{\midvalb}{(\minvalb+\maxvalb)/2}  % Midpoint value for the yellow transition

\newcommand{\colorrangeb}[1]{%
    \pgfmathsetmacro{\PercentColor}{100.0*(#1-\minvalb)/(\maxvalb-\minvalb)}%
    \ifdim #1 pt < \minvalb pt
        \cellcolor{goodred}{#1}%
    \else
        \ifdim #1 pt > \maxvalb pt
            \cellcolor{goodgreen}{#1}%
        \else
            \ifdim #1 pt < \midvalb pt
                \pgfmathsetmacro{\PercentLow}{200.0*(#1-\minvalb)/(\maxvalb-\minvalb)}%
                \xdef\PercentLow{\PercentLow}%
                \cellcolor{goodyellow!\PercentLow!goodred}{#1}%
            \else
                \pgfmathsetmacro{\PercentHigh}{200.0*(#1-\midvalb)/(\maxvalb-\minvalb)}%
                \xdef\PercentHigh{\PercentHigh}%
                \cellcolor{goodgreen!\PercentHigh!goodyellow}{#1}%
            \fi
        \fi
    \fi
}

\pgfmathsetmacro{\minvalc}{0}  % Minimum value in the table
\pgfmathsetmacro{\maxvalc}{55}  % Maximum value in the table
\pgfmathsetmacro{\midvalc}{(\minvalc+\maxvalc)/2}  % Midpoint value for the yellow transition

\newcommand{\colorrangec}[1]{%
    \pgfmathsetmacro{\PercentColor}{100.0*(#1-\minvalc)/(\maxvalc-\minvalc)}%
    \ifdim #1 pt < \minvalc pt
        \cellcolor{goodred}{#1}%
    \else
        \ifdim #1 pt > \maxvalc pt
            \cellcolor{goodgreen}{#1}%
        \else
            \ifdim #1 pt < \midvalc pt
                \pgfmathsetmacro{\PercentLow}{200.0*(#1-\minvalc)/(\maxvalc-\minvalc)}%
                \xdef\PercentLow{\PercentLow}%
                \cellcolor{goodyellow!\PercentLow!goodred}{#1}%
            \else
                \pgfmathsetmacro{\PercentHigh}{200.0*(#1-\midvalc)/(\maxvalc-\minvalc)}%
                \xdef\PercentHigh{\PercentHigh}%
                \cellcolor{goodgreen!\PercentHigh!goodyellow}{#1}%
            \fi
        \fi
    \fi
}

\pgfmathsetmacro{\minvald}{0}  % Minimum value in the table
\pgfmathsetmacro{\maxvald}{53}  % Maximum value in the table
\pgfmathsetmacro{\midvald}{(\minvald+\maxvald)/2}  % Midpoint value for the yellow transition

\newcommand{\colorranged}[1]{%
    \pgfmathsetmacro{\PercentColor}{100.0*(#1-\minvald)/(\maxvald-\minvald)}%
    \ifdim #1 pt < \minvald pt
        \cellcolor{goodred}{#1}%
    \else
        \ifdim #1 pt > \maxvald pt
            \cellcolor{goodgreen}{#1}%
        \else
            \ifdim #1 pt < \midvald pt
                \pgfmathsetmacro{\PercentLow}{200.0*(#1-\minvald)/(\maxvald-\minvald)}%
                \xdef\PercentLow{\PercentLow}%
                \cellcolor{goodyellow!\PercentLow!goodred}{#1}%
            \else
                \pgfmathsetmacro{\PercentHigh}{200.0*(#1-\midvald)/(\maxvald-\minvald)}%
                \xdef\PercentHigh{\PercentHigh}%
                \cellcolor{goodgreen!\PercentHigh!goodyellow}{#1}%
            \fi
        \fi
    \fi
}

\pgfmathsetmacro{\minvale}{0}  % Minimum value in the table
\pgfmathsetmacro{\maxvale}{84}  % Maximum value in the table
\pgfmathsetmacro{\midvale}{(\minvale+\maxvale)/2}  % Midpoint value for the yellow transition

\newcommand{\colorrangee}[1]{%
    \pgfmathsetmacro{\PercentColor}{100.0*(#1-\minvale)/(\maxvale-\minvale)}%
    \ifdim #1 pt < \minvale pt
        \cellcolor{goodred}{#1}%
    \else
        \ifdim #1 pt > \maxvale pt
            \cellcolor{goodgreen}{#1}%
        \else
            \ifdim #1 pt < \midvale pt
                \pgfmathsetmacro{\PercentLow}{200.0*(#1-\minvale)/(\maxvale-\minvale)}%
                \xdef\PercentLow{\PercentLow}%
                \cellcolor{goodyellow!\PercentLow!goodred}{#1}%
            \else
                \pgfmathsetmacro{\PercentHigh}{200.0*(#1-\midvale)/(\maxvale-\minvale)}%
                \xdef\PercentHigh{\PercentHigh}%
                \cellcolor{goodgreen!\PercentHigh!goodyellow}{#1}%
            \fi
        \fi
    \fi
}

\pgfmathsetmacro{\minvalz}{0}  % Minimum value in the table
\pgfmathsetmacro{\maxvalz}{49}  % Maximum value in the table
\pgfmathsetmacro{\midvalz}{(\minvalz+\maxvalz)/2}  % Midpoint value for the yellow transition

\newcommand{\colorrangez}[1]{%
    \pgfmathsetmacro{\PercentColor}{100.0*(#1-\minvalz)/(\maxvalz-\minvalz)}%
    \ifdim #1 pt < \minvalz pt
        \cellcolor{goodred}{#1}%
    \else
        \ifdim #1 pt > \maxvalz pt
            \cellcolor{goodgreen}{#1}%
        \else
            \ifdim #1 pt < \midvalz pt
                \pgfmathsetmacro{\PercentLow}{200.0*(#1-\minvalz)/(\maxvalz-\minvalz)}%
                \xdef\PercentLow{\PercentLow}%
                \cellcolor{goodyellow!\PercentLow!goodred}{#1}%
            \else
                \pgfmathsetmacro{\PercentHigh}{200.0*(#1-\midvalz)/(\maxvalz-\minvalz)}%
                \xdef\PercentHigh{\PercentHigh}%
                \cellcolor{goodgreen!\PercentHigh!goodyellow}{#1}%
            \fi
        \fi
    \fi
}

\pgfmathsetmacro{\minvalq}{0}  % Minimum value in the table
\pgfmathsetmacro{\maxvalq}{40}  % Maximum value in the table
\pgfmathsetmacro{\midvalq}{(\minvalq+\maxvalq)/2}  % Midpoint value for the yellow transition

\newcommand{\colorrangeq}[1]{%
    \pgfmathsetmacro{\PercentColor}{100.0*(#1-\minvalq)/(\maxvalq-\minvalq)}%
    \ifdim #1 pt < \minvalq pt
        \cellcolor{goodred}{#1}%
    \else
        \ifdim #1 pt > \maxvalq pt
            \cellcolor{goodgreen}{#1}%
        \else
            \ifdim #1 pt < \midvalq pt
                \pgfmathsetmacro{\PercentLow}{200.0*(#1-\minvalq)/(\maxvalq-\minvalq)}%
                \xdef\PercentLow{\PercentLow}%
                \cellcolor{goodyellow!\PercentLow!goodred}{#1}%
            \else
                \pgfmathsetmacro{\PercentHigh}{200.0*(#1-\midvalq)/(\maxvalq-\minvalq)}%
                \xdef\PercentHigh{\PercentHigh}%
                \cellcolor{goodgreen!\PercentHigh!goodyellow}{#1}%
            \fi
        \fi
    \fi
}

\pgfmathsetmacro{\minvalr}{0}  % Minimum value in the table
\pgfmathsetmacro{\maxvalr}{42}  % Maximum value in the table
\pgfmathsetmacro{\midvalr}{(\minvalr+\maxvalr)/2}  % Midpoint value for the yellow transition

\newcommand{\colorranger}[1]{%
    \pgfmathsetmacro{\PercentColor}{100.0*(#1-\minvalr)/(\maxvalr-\minvalr)}%
    \ifdim #1 pt < \minvalr pt
        \cellcolor{goodred}{#1}%
    \else
        \ifdim #1 pt > \maxvalr pt
            \cellcolor{goodgreen}{#1}%
        \else
            \ifdim #1 pt < \midvalr pt
                \pgfmathsetmacro{\PercentLow}{200.0*(#1-\minvalr)/(\maxvalr-\minvalr)}%
                \xdef\PercentLow{\PercentLow}%
                \cellcolor{goodyellow!\PercentLow!goodred}{#1}%
            \else
                \pgfmathsetmacro{\PercentHigh}{200.0*(#1-\midvalr)/(\maxvalr-\minvalr)}%
                \xdef\PercentHigh{\PercentHigh}%
                \cellcolor{goodgreen!\PercentHigh!goodyellow}{#1}%
            \fi
        \fi
    \fi
}

\pgfmathsetmacro{\minvals}{0}  % Minimum value in the table
\pgfmathsetmacro{\maxvals}{30}  % Maximum value in the table
\pgfmathsetmacro{\midvals}{(\minvals+\maxvals)/2}  % Midpoint value for the yellow transition

\newcommand{\colorranges}[1]{%
    \pgfmathsetmacro{\PercentColor}{100.0*(#1-\minvals)/(\maxvals-\minvals)}%
    \ifdim #1 pt < \minvals pt
        \cellcolor{goodred}{#1}%
    \else
        \ifdim #1 pt > \maxvals pt
            \cellcolor{goodgreen}{#1}%
        \else
            \ifdim #1 pt < \midvals pt
                \pgfmathsetmacro{\PercentLow}{200.0*(#1-\minvals)/(\maxvals-\minvals)}%
                \xdef\PercentLow{\PercentLow}%
                \cellcolor{goodyellow!\PercentLow!goodred}{#1}%
            \else
                \pgfmathsetmacro{\PercentHigh}{200.0*(#1-\midvals)/(\maxvals-\minvals)}%
                \xdef\PercentHigh{\PercentHigh}%
                \cellcolor{goodgreen!\PercentHigh!goodyellow}{#1}%
            \fi
        \fi
    \fi
}

\definecolor{cvprblue}{rgb}{0.21,0.49,0.74}
\usepackage[pagebackref,breaklinks,colorlinks,allcolors=cvprblue]{hyperref}

\def\paperID{975} % *** Enter the Paper ID here
\def\confName{CVPR}
\def\confYear{2025}

\title{VoxTell: Free-Text Promptable Universal 3D Medical Image Segmentation}

\author{Maximilian Rokuss\textsuperscript{1,2,6}\thanks{\small Contributed equally. Each co-first author may list themselves as lead author on their CV.}\;,
Moritz Langenberg\textsuperscript{1,2}\footnotemark[1]\;,
Yannick Kirchhoff\textsuperscript{1,2,6},
Fabian Isensee\textsuperscript{1,4},\\
Benjamin Hamm\textsuperscript{1,3},
Constantin Ulrich\textsuperscript{1,3},
Sebastian Regnery\textsuperscript{5},
Lukas Bauer\textsuperscript{5},\\
Efthimios Katsigiannopulos\textsuperscript{5},
Tobias Norajitra\textsuperscript{1,7},
Klaus Maier-Hein\textsuperscript{1,2,3,4,6,7}
\\
\and
\textsuperscript{1}German Cancer Research Center, Division of Medical Image Computing, Germany\\
\textsuperscript{2}Faculty of Mathematics and Computer Science and 
\textsuperscript{3}Medical Faculty  - Heidelberg University\\
\textsuperscript{4}Helmholtz Imaging, 
\textsuperscript{5}Department of Radiation Oncology, Heidelberg University Hospital, Germany\\
\textsuperscript{6}HIDSS4Health, Heidelberg
\textsuperscript{7}Pattern Analysis and Learning Group, Heidelberg University Hospital
\\
{\tt\small \{maximilian.rokuss,  moritz.langenberg\}@dkfz-heidelberg.de}
\and
}

\begin{document}
%\maketitle

\twocolumn[{
\renewcommand\twocolumn[1][]{#1}%
\maketitle
\vspace{-2em}
\centering
\includegraphics[width=\textwidth]{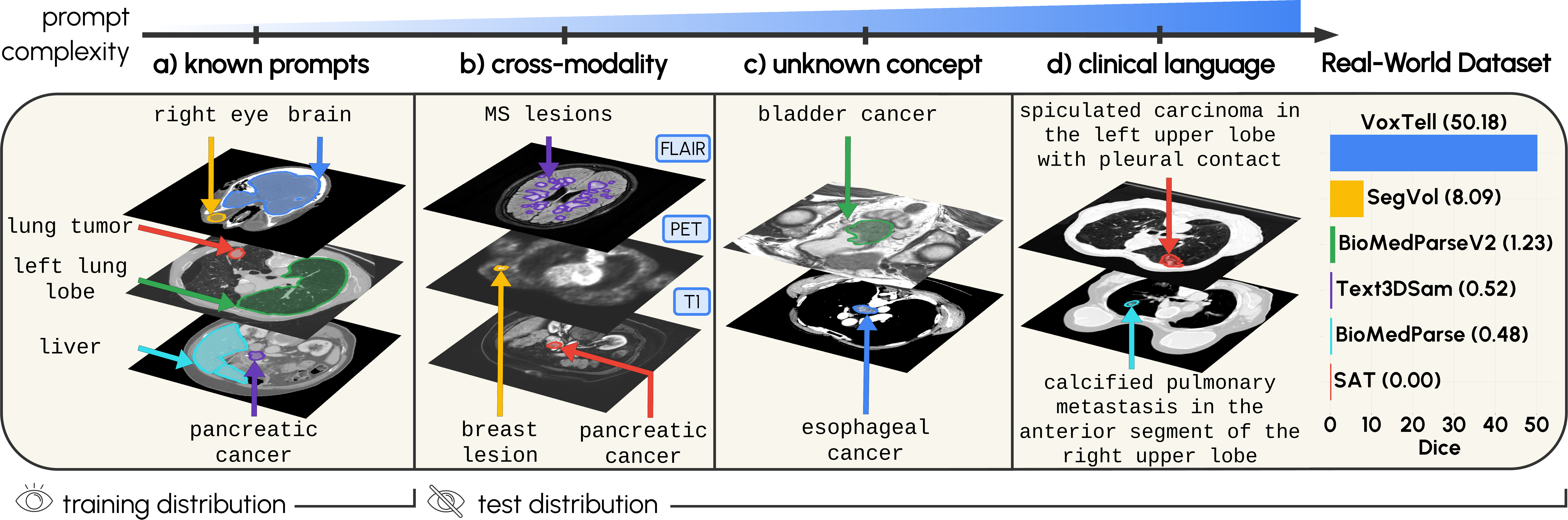}
\captionsetup{hypcap=false}  % Disable hypcap for this particular caption
\captionof{figure}{
\textbf{VoxTell performs 3D medical image segmentation directly from arbitrary free-text prompts.} The figure shows progressively challenging scenarios: (a) known anatomical structures seen during training, (b) generalization of learned concepts to other imaging modalities, (c) novel concepts never encountered during training, and (d) clinical language understanding from real radiology reports with spatially grounded descriptions. The bar chart (right) reports Dice scores on a held-out radiotherapy cohort using report-derived prompts, shown in (d), where VoxTell outperforms prior text-promptable segmentation methods. 
\vspace{1em}
}
\label{fig:teaser}
}]
\footnotetext[1]{Contributed equally. Co-first authors may list as lead on CV.}

\begin{abstract}
We introduce VoxTell, a vision–language model for text-prompted volumetric medical image segmentation. It maps free-form descriptions, from single words to full clinical sentences, to 3D masks. Trained on 62K+ CT, MRI, and PET volumes spanning 1K+ anatomical and pathological classes, VoxTell uses multi-stage vision–language fusion across decoder layers to align textual and visual features at multiple scales. It achieves state-of-the-art zero-shot performance across modalities on unseen datasets, excelling on familiar concepts while generalizing to related unseen classes. Extensive experiments further demonstrate strong cross-modality transfer, robustness to linguistic variations and clinical language, as well as accurate instance-specific segmentation from real-world text. Code is available at: \url{www.github.com/MIC-DKFZ/VoxTell}
\end{abstract}    
\section{Introduction} %  like “calcified nodules in right lung parenchyma”
\label{sec:intro}

Accurate segmentation of organs and pathologies in volumetric medical images is a cornerstone of modern diagnosis and treatment planning. Despite substantial progress, most methods remain specialized to particular structures or modalities, and even models capable of segmenting a multitude of categories fail to generalize beyond their training distribution. This fragmentation has driven the development of general-purpose segmentation frameworks, such as \textit{segment anything} paradigms aiming for universal, task-agnostic segmentation. SAM~\cite{sam1} demonstrates that a single model can generalize across diverse natural images using spatial prompts. Early medical adaptations, including MedSAM~\cite{medsam} and subsequent approaches~\cite{3dsamadapter,cheng2023sammed2d,roy2023sammd,segvol,vista3d,prism,sammed3d,scribbleprompt,lesionlocator,nninteractive} present large-scale foundation models for medical images, enabling universal segmentation. While promising, they rely on manual interactive prompts such as points, bounding boxes, or scribbles to specify target structures.\\

\noindent Beyond interactive spatial prompts, text-based segmentation offers a natural interface for clinicians, allowing direct description of anatomical or pathological structures or leveraging existing radiology reports, reducing manual input. Text prompts further open the possibility of exploiting the rich, structured embeddings of modern language models, which encode semantic knowledge of anatomical concepts and inter-structure relationships. Leveraging these embeddings could, in principle, guide segmentation models to generalize to related but unseen structures or modalities, a largely unexplored capability. 
In the natural-image domain, recent works have successfully integrated language queries with segmentation architectures, enabling open-vocabulary segmentation from free-form textual prompts~\cite{lai2024lisa,rasheed2024glammpixelgroundinglarge,huang2025DETRIS,wu2023generalobjectfoundationmodel,liu2024UniLSeg,zhang2024evfsamearlyvisionlanguagefusion,xie2025RICE,wei2024instructseg,wei2024hyperseguniversalvisualsegmentation,zhang2024psalm,chen2024sam4mllm,an2024MLCD-seg,anonymous2025sam3,wang2025xsam}. These advances demonstrate that text-driven prompts can offer a flexible complement to spatial interactions.

\noindent Existing text-guided medical segmentation models~\cite{biomedparse,boltzformer,segvol,SAT} have made notable progress, yet they largely behave like multitask segmentation networks trained on fixed concepts or predefined ontologies. In addition, they are sensitive to phrasing, synonyms, and minor spelling variations. Consequently, their ability to handle unseen or sentence-level prompts is limited, offering little benefit over models such as TotalSegmentator~\cite{wasserthal2023totalsegmentator} trained on predefined categories. Performance typically degrades when moving from simple single-word labels (“liver”) to complex, clinically descriptive queries, precisely the scenarios where flexible language understanding would be most valuable. Medical text is highly specialized, often describing instance-specific spatial and semantic relationships (\eg “calcified nodules in right lung parenchyma”), rather than fixed categories. A key promise of text-promptable models is their potential to generalize more broadly, using well-structured language embeddings to extrapolate to related but unseen concepts. Our work takes a step in this direction by using repeated vision–language fusion, moving \textit{towards} open-set 3D segmentation.\\

\noindent We introduce \textit{VoxTell}, a 3D vision–language segmentation model that maps free-form textual queries to volumetric masks. VoxTell is trained on a large-scale 3D dataset spanning CT, PET, and multiple MRI modalities, incorporating a pretrained text encoder to robustly interpret diverse linguistic inputs, from single words to longer sentences, including synonyms and minor spelling errors. Unlike many prior models that perform only late-stage fusion between text and visual features, VoxTell employs a multi-stage fusion strategy throughout the decoder hierarchy. This repeated cross-modal interaction enables continuous alignment between textual prompts and volumetric features, which is particularly effective for instance-specific or descriptive clinical queries.\\

\noindent Extensive experiments demonstrate that VoxTell achieves state-of-the-art performance across known categories and shows promising generalization to semantically related unseen concepts, all on held-out datasets. It maintains robustness under varied textual formulations, an area where prior models often struggle, particularly for clinically descriptive or instance-specific prompts. These results highlight its potential for flexible, text-promptable 3D segmentation. In summary, our key contributions are as follows:

\begin{itemize}
    \item \textbf{VoxTell:} A text-prompted 3D segmentation model that directly maps free-form clinical prompts, including sentence-level clinical descriptions, to volumetric masks.
    \item \textbf{Multi-stage vision–language fusion:} Repeated cross-modal interactions at multiple decoder stages improve alignment between textual and volumetric features.
    \item \textbf{State-of-the-art text-promptable segmentation:} Validated across multiple structures and modalities (CT, PET, MRI), VoxTell outperforms prior text-guided segmentation models.
    \item \textbf{Towards open-set generalization:} Effective across modality shifts and unseen related structures.
\end{itemize}

% In the natural-image domain, models such as 
% gerad in Progress!: 
% LISA~\cite{lai2024lisa}, GLaMM~\cite{rasheed2024glammpixelgroundinglarge} (nutzen SAM mask decoder (im SAM paper dann auch wieder "inspired by MaskFormer")). 
% DETRIS~\cite{huang2025DETRIS} (auch mask prediction)
% GLEE~\cite{wu2023generalobjectfoundationmodel} (also mask prediction) 
% UniLSeg~\cite{liu2024UniLSeg} (also mask prediction)
% EVF-SAM~\cite{zhang2024evfsamearlyvisionlanguagefusion} (SAM extension)
% RICE~\cite{xie2025RICE} (region attention stuff)
% InstructSeg~\cite{wei2024instructseg}, HyperSeg~\cite{wei2024hyperseguniversalvisualsegmentation} ,and 
% PSALM~\cite{zhang2024psalm} extend MaskFormer-based architectures with language queries, towards enabling open-vocabulary segmentation from free-form textual prompts.  SAM4MLLM~\cite{chen2024sam4mllm}, 
% MLCD-seg~\cite{an2024MLCD-seg} (cluster stuff which is not interesting for us.)
% SAM3~\cite{anonymous2025sam3} (SAM 3: Segment Anything with Concepts)
% X-SAM~\cite{wang2025xsam}. These advances indicate that text-driven prompts may offer a flexible complement to spatial interactions.\\

\section{Text-Guided Medical Image Segmentation}
\label{sec:related_work}

% \begin{figure*}[t]
%   \centering
%   \includegraphics[width=\textwidth]{figures/fig2_v3-cropped.pdf}
%   \caption{\textbf{Overview of VoxTell.} \textit{Left:} high-level pipeline showing the 3D UNet encoder, text embedding, and decoder with multi-stage text-guided fusion. \textit{Right:} schematic of repeated vision–language fusion at each decoder stage with deep supervision. VoxTell generalizes the extends the MaskFormer principle of query–image dot-product fusion from a single late stage to multi-scale, repeated interactions, enabling continuous alignment between textual prompts and volumetric features.}
%   \label{fig:fig1}
% \end{figure*}

% feedback plz
\begin{figure*}[t]
  \centering
  \includegraphics[width=\textwidth]{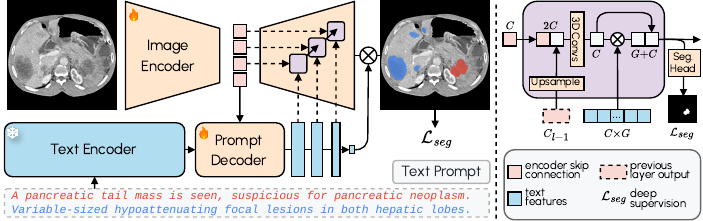}
  \caption{\textbf{Overview of VoxTell.} \textit{Left:} A 3D image volume is encoded into latents, while a free-text prompt is first embedded and then processed by a prompt decoder to produce multi-scale text features that guide image decoding. \textit{Right:} The decoder performs multi-stage vision–language fusion: at each resolution, text embeddings modulate volumetric features, extending MaskFormer-style query–image fusion to multiple scales with deep supervision.}
  \label{fig:fig1}
  \vspace{-1em}
\end{figure*}

Recent advances in medical image segmentation have explored the integration of textual guidance with visual models. They can be categorized into three main paradigms:

\paragraph{Closed-set segmentation via head mapping} In this line of work, models are trained on a fixed set of output classes or concepts, with textual prompts used to select or retrieve the appropriate segmentation head. This approach was pioneered by the CLIP-driven Universal Model~\cite{liu2023clip}, which achieves strong in-domain performance, but cannot segment arbitrary structures from text. This is similar to classical multi-dataset segmentation networks, such as MultiTalent~\cite{multitalent}, which learn multiple concepts simultaneously but remain limited to predefined labels. Building on this paradigm, CAT~\cite{CAT} coordinates anatomical and textual prompts to refine segmentation queries, achieving strong organ and tumor segmentation; however, it still relies on training-set categories and cannot fully generalize to unseen structures via text alone.\\

\paragraph{Text as auxiliary guidance} Another approach leverages textual descriptions as supplementary supervision for per-dataset, class-specific segmentation, improving performance over conventional fully supervised networks~\cite{LViT,Textmatch_MICCAI24}. These methods typically focus on benchmarks such as QaTa-COV19~\cite{Degerli_2022} and MosMed++~\cite{mosmeddata} and do not explicitly address generalization to unseen categories. R-Super~\cite{r-super} extends this idea by converting entire radiology reports into auxiliary supervision for segmentation. However, such methods remain dataset-specific or cannot segment structures given an unseen text prompt.

\paragraph{Text-promptable segmentation models}
The most ambitious line of work aims to produce segmentation masks directly from arbitrary text prompts. Early 2D models like BioMedParse~\cite{biomedparse} were trained on millions of biomedical image–mask–text triplets, covering 64 categories across nine modalities, building on a Mask2Former-style architecture~\cite{mask2former} where text and image features are fused at a late decoding stage via cross-attention. BoltzFormer~\cite{boltzformer} (BiomedParseV2) further improves small-object segmentation with Boltzmann-distributed attention. In 3D, SegVol~\cite{segvol} pretrains on 96K CT volumes and fine-tunes on 6K labeled images, supporting point, box, and text prompts for over 200 anatomical classes, while Text3DSAM~\cite{xin2025text3dsam} adapts SAM for 3D text-guided segmentation. SAT~\cite{SAT} proposes a medical-domain-informed text encoder with a large multi-center corpus (22K scans, 497 classes), achieving specialist-level performance, particularly on rare tail categories. It follows a MaskFormer~\cite{maskformer} paradigm, fusing textual embeddings with high-resolution visual features via late-stage dot product.\\

\noindent Despite impressive progress, existing text-guided segmentation models still fall short of the main promise of language-driven segmentation: leveraging structured linguistic knowledge to delineate diverse, clinically meaningful structures. Most methods 1) rely on fixed label sets, 2) generalize poorly to free-form phrasing or descriptive queries, and 3) lack evaluation on unseen concepts or modalities. As a result, they function more like closed-set segmentation models that use text merely to select a predefined mask, rather than models capable of parsing arbitrary descriptions.
\noindent VoxTell addresses these limitations through three key design choices: training on a large, diverse 3D dataset spanning CT, PET, and MRI; extensive vocabulary expansion to handle prompt complexity complemented by benchmarking for a SoTA pretrained text encoder; and introducing a multi-stage vision–language fusion strategy that injects text guidance throughout the decoder hierarchy. Prior MaskFormer-based models~\cite{SAT,lai2024lisa,wei2024instructseg,zhang2024psalm} perform a single late-stage fusion between text and high-resolution image features, limiting responsiveness to complex prompts. VoxTell extends this approach to repeated fusion across all decoder stages, enabling continuous, multi-scale interaction between textual and volumetric features.
\section{Multi-Stage Vision-Language Fusion}
\label{sec:method}

VoxTell is a 3D vision–language segmentation model that generates volumetric masks from free-form text prompts, illustrated in Fig. \ref{fig:fig1}. Similar to prior approaches our method is inspired by MaskFormer, but introduces key modifications to enable effective cross-modal interaction. While MaskFormer, SAT, and Mask2Former~\cite{mask2former} use image bottleneck or decoder features as memory for the transformer decoder, \textit{the final segmentation is obtained via a single dot product between query features and generic image features}. This late fusion forces the image backbone to learn unspecific, prompt-agnostic representations, receiving no guidance from the query until the last output step. % As a result, responsiveness to prompts is limited and intermediate features remain under-specialized.\\

\noindent We argue that robust free-text-promptable segmentation in 3D requires \textit{repeated cross-modal interaction} throughout the decoding hierarchy. Location-specific prompts such as “lesion in right lung” are difficult to capture with a single, shared segmentation head sliding across the volume, since it cannot adapt to spatially grounded queries. To address this, VoxTell injects textual embeddings into multi-scale decoder features at multiple depths, aligning linguistic and spatial information throughout the decoder network.\\

\noindent While transformers dominate many 2D segmentation models, we deliberately adopt a UNet-style backbone~\cite{isensee_nnu-net_2021,mednext} for volumetric encoding. UNet-based architectures remain state of the art in 3D medical imaging, consistently outperforming transformer counterparts across large-scale benchmarks~\cite{bassi2024touchstonebenchmark,nnunet_revisited} and recent challenges~\cite{uls_challenge,topcowchallenge,autopet_winner}. VoxTell integrates text conditioning directly into UNet’s multi-scale feature maps, ensuring that prompt information influences intermediate representations rather than being applied only at the output stage.\\

\noindent Finally, this design can leverage \textit{deep supervision} across decoder stages: By enforcing prompt-conditioned predictions at multiple resolutions, the model is compelled to incorporate guidance from text early in the feature hierarchy, resulting in masks that better adhere to the input query. We now formalize this architecture in mathematical detail:

\paragraph{Overview}
Let $V \in \mathbb{R}^{H \times W \times D}$ be a 3D input volume. A UNet-style encoder $f_\text{enc}$ extracts multi-scale feature maps:
\begin{equation}
\mathcal{Z} = \{z_1, \dots, z_S\}, \quad z_s \in \mathbb{R}^{C_s \times H_s \times W_s \times D_s},
\end{equation}
where $S$ is the number of scales. A free-form textual prompt $p$ is embedded via a pretrained text encoder $f_\text{text}$:
\begin{equation}
q = f_\text{text}(p) \in \mathbb{R}^{d}.
\end{equation}
\noindent A transformer-based prompt decoder $f_{\text{prompt}}$ takes $q$ as query and bottleneck features $z_S$ as key--value pairs. Its output is projected through stage-specific adapter MLPs to produce multi-scale text-guidance tensors
\begin{equation}
\mathcal{T} = f_\text{prompt}(q, z_S) = \{T_1, \dots, T_S\}, \quad T_s \in \mathbb{R}^{G \times C_s},
\end{equation}
where each $T_s$ aligns with the corresponding decoder feature dimension, and $G$ denotes the guidance embedding dimension (set to $32$), yielding scale-specific textual modulation.

\paragraph{Cross-Scale Fusion}
The UNet-style decoder $f_\text{dec}$ reconstructs features from coarse to fine resolution, fusing image features $\mathcal{Z}$ with textual guidance $\mathcal{T}$ at every stage. At each scale $s$, the upsampled output from the previous stage $y_{s-1}^\uparrow$ is first concatenated with the encoder skip connection $z_s$ and passed through convolutional blocks:
\begin{equation}
z'_s = \text{ConvBlock}(\text{concat}(y_{s-1}^\uparrow, z_s)).
\end{equation}

\noindent To inject prompt conditioning, we extend the MaskFormer principle of using dot products between image and text features to \emph{all scales}, applying a channel-wise dot product between $z'_s$ and $T_s$, which is then concatenated back to the intermediate feature:
\begin{equation}
y_s = \text{concat}\big(z'_s, T_s \odot z'_s\big), \quad y_s \in \mathbb{R}^{(C_s+G) \times H_s \times W_s \times D_s},
\end{equation}
where $\odot$ denotes a dot product along $C_s$, resulting in $G$ new channels. The stage-wise update is then
\begin{equation}
y_{s} = f_{\text{dec},s}(y_{s-1}, z_s, T_s).
\end{equation}

\paragraph{Deep Supervision}
Each intermediate decoder output $y_s$ is further mapped to a prediction $\hat{Y}_s$ by a segmentation head, enabling auxiliary supervision across scales. The overall loss is defined as

\begin{equation}
\mathcal{L} = \sum_{s=1}^{S} \lambda_s \, \mathcal{L}_\text{seg}(\hat{Y}_s, Y_s),
\end{equation}

where $Y_s$ is the ground-truth mask downsampled to scale $s$, $\mathcal{L}_\text{seg}$ combines Dice and cross-entropy losses, and $\lambda_s$ are scale weights. Deep supervision promotes early integration of textual features, compelling the initial decoder stages to incorporate the textual queries.\\

\noindent In summary, VoxTell performs \emph{multi-scale, repeated vision–language fusion}, complemented with deep supervision. By modulating decoder features with prompt embeddings at every resolution, the model learns conditioned volumetric representations that improve segmentation accuracy and remain robust to diverse free-text queries.

\section{Dataset and Vocabulary Construction}
\label{sec:data}

\paragraph{Large-Scale Multi-Modality Dataset} To train and evaluate VoxTell, we curated a large-scale, multi-modality 3D medical imaging corpus comprising 158 publicly available datasets with over 62K volumetric scans ($\approx$4~TB) and 1,087 anatomical and pathological concepts. The collection spans Computed Tomography (CT), multi-sequence Magnetic Resonance Imaging (MRI), and Positron Emission Tomography (PET), encompassing both healthy and pathological anatomies. In scale and diversity, it more than doubles the number of datasets and nearly triples the number of volumes compared to the largest prior compilation~\cite{SAT}. Label sets range from large organ delineations to fine-grained substructures and lesions, providing rich semantic diversity for learning language-conditioned representations. Full dataset details and access links are listed in Appendix~\ref{sec:dataset_details}. For model development and ablation studies, 5\% of each dataset is reserved for validation to ensure test integrity, and all ablations are conducted on this split.\\

\paragraph{Comprehensive Vocabulary Construction} We aim to ensure broad textual coverage of diverse medical concepts across heterogeneous datasets. First, we harmonize label semantics by unifying synonymous or overlapping class names, resolving ambiguities where identical terms differ across datasets (\eg whether "liver" includes lesions). Next, we expand the label space using dataset metadata to generate anatomically precise variants (\eg "right kidney" $\rightarrow$ "right renal organ", "dexter kidney") and compose hierarchical aggregates (combining "left rib 1"–"left rib 12" into "left rib cage") via a large language model. The full semantic standardization pipeline, including label expansion, cross-dataset harmonization, and validation, is detailed in Appendix~\ref{sec:appendix_agentic_system}. The final vocabulary comprises 1,087 unified concepts and 9,682 rewritten labels, sampled during training with emphasis on the main term (\eg "liver" rather than "hepatic organ").\\

\paragraph{Instance-Focused Dataset} To handle fine-grained, spatially grounded prompts, \ie localizing individual instances within a semantic class, we complement our semantic dataset with an instance-focused dataset, assembled from publicly available sources. This dataset is used for the ReXGroundingCT~\cite{rexgroundingct} benchmark, which provides instance-level annotations. The dataset combines (i) the official ReXGroundingCT training split linking free-text report findings to precise 3D segmentations, (ii) semantic lesion datasets converted to instance-level annotations using TotalSegmentator, and (iii) TCIA collections with structured DICOM metadata. Full construction details are provided in Appendix~\ref{sec:appendix_instance_dataset_creation}. For evaluation on ReXGroundingCT and to ensure a fair comparison, VoxTell and competing methods are fine-tuned on this dataset to enable assessment of instance-specific localization.
\section{Experiments}
\label{sec:experiments}

We evaluate VoxTell along four key axes: (i) standard anatomical and pathological segmentation, (ii) robustness to diverse text prompts, (iii) generalization to unseen concepts and cross-modality scenarios, and (iv) clinical language understanding and instance-level segmentation from reports.

\paragraph{From In-Distribution to Out-of-Distribution Benchmarks}
A common practice among text-promptable baselines is to benchmark on the same dataset and anatomical structures used for training via a standard train/test split~\cite{SAT}, assessing performance on in-distribution images and concepts. In contrast, we adopt a more challenging and realistic setting by evaluating exclusively on unseen datasets, \ie out-of-distribution (OOD) images. Our experiments include both familiar and novel concepts, as well as familiar classes in new modalities, providing a rigorous assessment of generalization in open-world scenarios.

\paragraph{Benchmarking on Clinical Language} To ensure clinical reliability, we benchmark exclusively on expert-curated, manually annotated datasets. We use the public ReXGroundingCT~\cite{rexgroundingct} benchmark with official splits. We further assemble an in-house cohort of 203 patients undergoing stereotactic body radiotherapy (SBRT) for primary or secondary lung tumors: Each patient has planning CT scans with expert-annotated gross tumor volume (GTV) contours. Corresponding textual descriptions were extracted from radiology reports, exemplified in Fig.~\ref{fig:teaser}d. This dataset was held out entirely during training. Details in Appendix \ref{sec:in_house_data}.

\paragraph{Ablation Studies}
We systematically investigate the impact of architectural choices: (i) single-stage versus multi-stage fusion, (ii) inclusion of deep supervision across decoder stages, and (iii) training batch size scaling. These ablations isolate the contributions of repeated text-image interactions and intermediate supervision to VoxTell’s performance.

\paragraph{Evaluation Metrics}
Segmentation quality is measured using the Dice coefficient. For instance-level evaluation on ReXGroundingCT, we additionally report HIT$_{5\%}$, the fraction of samples achieving Dice $\ge 5\%$. To quantify text robustness, we analyze performance across prompt variations including synonyms, rephrasings, and minor typographical errors.

\paragraph{Implementation Details}
VoxTell integrates a ResEncL~\cite{nnunet_revisited} vision backbone with six encoder stages. Text prompts are embedded using the frozen Qwen3-Embedding-4B~\cite{qwen3embedding}, selected based on ablation results in Appendix~\ref{sec:appendix_text_encoder_selection}. The prompt decoder is a six-layer transformer with a 2048-dimensional query space. Multi-scale vision–language feature fusion is applied at all decoder stages with deep supervision. Experiments are implemented in PyTorch. Ablation studies use a single NVIDIA A100 GPU (batch size 2), while the final model is trained on 64 A100 GPUs with a batch size of 128, requiring roughly six days. Optimization uses SGD with an initial learning rate of $1\times10^{-4}$ and polynomial decay. During training, both positive and negative prompts are sampled, \ie prompts not present in the image. Further details are provided in Appendix~\ref{sec:appendix_implementation_details}.
\section{Results and Discussion}
\label{sec:results}

\begin{table*}[h!]
\centering
\setlength{\tabcolsep}{2pt}
\begin{adjustbox}{max width=\textwidth}
\begin{tabular}{l|ccccc|cccccc|c}
\toprule
Dataset & \shortstack{Abd. \\ Organs} & \shortstack{Lung \&\\ Airway} & \shortstack{Liver \&\\ Vessels} & \shortstack{Head \&\\ Neck} & Heart & \shortstack{Lung \\ Tumor} & \shortstack{Multiple \\ Sclerosis} & \shortstack{Adrenal \\ Tumor} & \shortstack{Liver \\ Lesions} & \shortstack{Bones w. \\ Fractures} & \shortstack{Brain \\ Mets} & \shortstack{Mean \\ Dice} \\
         & \cite{pediatric-ct-seg} & \cite{2023aeropath} & \cite{toprak2025veela} & \cite{radcure} & \cite{pace2024hvsmr} & \cite{kinahan2019acrin-nsclc} & \cite{isbi-ms} & \cite{adrenalacc} & \cite{hcctace} & \cite{pengwin} & \cite{stanfordBrainMets} &  \\
Modality & \CT & \CT & \CT & \CT & \MRI & \PET & \MRI & \CT & \CT & \CT & \MRI &  \\
\midrule
TotalSegmentator~\cite{wasserthal2023totalsegmentator} & \colorrange{67.64} & \colorrangea{85.23} & \colorrangeb{49.48} & - & - & - & - & - & - & - & - & - \\
BioMedParse~\cite{biomedparse} & \colorrange{9.12} & \colorrangea{0.00} & \colorrangeb{6.53} & \colorrangec{1.66} & \colorranged{8.12} & \colorrangee{2.73}\textsuperscript{\textbf{*}} & \colorrange{9.57} & \colorrange{52.91} & \colorrange{41.25} & \colorrange{0.47} & \colorrangez{9.61} & 12.91 \\
Text3DSam~\cite{xin2025text3dsam} & \colorrange{26.56} & \colorrangea{65.36} & \colorrangeb{28.91} & \colorrangec{1.34} & \colorranged{15.65} & \colorrangee{12.21} & \colorrange{0.21} & \colorrange{0.00} & \colorrange{0.52} & \colorrange{24.75} & \colorrangez{0.10} & 15.97 \\
SegVol~\cite{segvol} & \colorrange{52.50} & \colorrangea{88.67} & \colorrangeb{41.34} & \colorrangec{17.93} & \colorranged{1.85}\textsuperscript{\textbf{*}} & \colorrangee{0.00}\textsuperscript{\textbf{*}} & \colorrange{0.00}\textsuperscript{\textbf{* \textdagger}} & \colorrange{51.08} & \colorrange{58.35} & \colorrange{22.11} & \colorrangez{0.00}\textsuperscript{\textbf{* \textdagger}} & 30.35 \\
BioMedParseV2~\cite{boltzformer} & \colorrange{51.78} & \colorrangea{62.59} & \colorrangeb{35.06} & \colorrangec{20.65} & \colorranged{18.11} & \colorrangee{0.47} & \colorrange{2.03} & \colorrange{55.86} & \colorrange{70.37} & \colorrange{1.35} & \colorrangez{18.66} & 30.59 \\
SAT~\cite{SAT} & \colorrange{68.79} & \colorrangea{87.98} & \colorrangeb{37.65} & \textbf{\colorrangec{54.33}} & \colorranged{43.38} & \colorrangee{77.13} & \colorrange{13.68} & \colorrange{0.09}\textsuperscript{\textbf{\textdagger}} & \colorrange{62.28} & \colorrange{96.05} & \colorrangez{22.16} & 51.23 \\
\midrule
\textbf{VoxTell (Ours)} & \textbf{\colorrange{72.94}} & \textbf{\colorrangea{89.65}} & \textbf{\colorrangeb{60.56}} & \colorrangec{51.28} & \textbf{\colorranged{52.72}} & \textbf{\colorrangee{83.24}} & \textbf{\colorrange{72.71}} & \textbf{\colorrange{77.23}} & \textbf{\colorrange{73.24}} & \textbf{\colorrange{97.59}} & \textbf{\colorrangez{48.19}} & \textbf{70.85} \\
\bottomrule
\end{tabular}
\end{adjustbox}
\caption{\textbf{Zero-shot segmentation performance (Dice)} on common anatomical (left) and pathological (right) concepts across unseen CT, MRI, and PET datasets. VoxTell surpasses classical (TotalSegmentator) and text-promptable baselines, with clear gains over the next-best method (SAT). Some methods entirely fail on certain cases. Models not trained on a modality (\textbf{*}) or pathology (\textbf{\textdagger}) are marked for clarity.}
\label{tab:model_comparison}
\end{table*}

We first evaluate VoxTell against state-of-the-art text-promptable segmentation methods on zero-shot 3D segmentation across unseen CT, MRI, and PET datasets, covering a wide range of anatomical structures and pathologies. We perform ablations to isolate the impact of multi-stage fusion and deep supervision, and demonstrate robustness to diverse textual prompt variations. Finally, we show VoxTell’s ability to generalize to known categories on new modalities as well as unseen structures, and perform precise instance-specific segmentation from real-world sentence-level findings, highlighting its versatility and clinical applicability.

\paragraph{State-of-the-art performance across anatomical and pathological structures} We evaluate VoxTell in a zero-shot setting on unseen 3D datasets spanning CT, MRI, and PET, including five healthy and six pathological datasets (Tab.~\ref{tab:model_comparison}). The benchmark covers both established structures, such as major organs and liver tumors, and rarer pathological concepts like adrenal tumors, multiple sclerosis lesions, and fractured bones. We indicate which baselines were not trained on certain modalities (\textbf{*}) or pathologies (\textbf{\textdagger}), allowing for a fairer interpretation of the results. Where applicable, we include TotalSegmentator, a classical autosegmentation tool without text prompts. We use the same standard anatomical and pathological terms as text prompts for all methods, without any prompt optimization or dataset-specific tuning. The full list of prompts, along with detailed per-class results, is provided in Appendix Tab.~\ref{tab:full_results_appendix}.\\

\noindent On structures seen by all methods during training (\eg abdominal organs and liver tumors), VoxTell outperforms both text-promptable baselines and TotalSegmentator, achieving the highest Dice on organs, with 89.7 on lung and airway, as well as 73.2 on liver lesions. Notably, the abdominal organ dataset~\cite{pediatric-ct-seg} consists of challenging pediatric CT scans, introducing a domain shift that explains the performance drop of TotalSegmentator and the lower Dice scores compared to adult CT. Furthermore, VoxTell consistently achieves the highest Dice scores across all pathological concepts. Baselines that were trained on the same pathologies still underperform, while those without exposure often fail entirely. This underscores VoxTell’s strength as a vision–language foundation model capable of segmenting a broad range of clinical targets. 

\begin{table}[t]
\centering
\setlength{\tabcolsep}{4pt}
\renewcommand{\arraystretch}{1.05}
\begin{adjustbox}{max width=\columnwidth}
\begin{tabular}{lccc}
\toprule
\textbf{Model} & \textbf{Fusion Stages} & \textbf{Deep Sup.} & \textbf{Dice} \\
\midrule
\multicolumn{4}{c}{\textit{Single-stage fusion (late only)}} \\
Mask2Former~\cite{maskformer} & 1 & \xmark & 51.68 \\
MaskFormer (SAT~\cite{SAT}) & 1 & \xmark & 55.11 \\
\midrule
\multicolumn{4}{c}{\textit{Multi-stage fusion (ours)}} \\
Ours (3 stages) & 3 & \xmark & 60.16 \\
Ours (5 stages) & 5 & \xmark & 61.54 \\
Ours (+ deep sup.) & 5 & \cmark & 62.55 \\
\midrule
\rowcolor{gray!10}
Ours (+ scaling) & 5 & \cmark & \textbf{69.43} \\
\bottomrule
\end{tabular}
\end{adjustbox}
\caption{\textbf{Ablation Study on Text–Image Fusion.}
Validation set results using identical training data and text encoder across all methods. We compare single-stage late fusion baselines (3D Mask2Former and MaskFormer as in SAT~\cite{SAT}) with our multi-stage fusion design. Increasing fusion stages and adding deep supervision progressively improve Dice performance. Note that fusion stages = 1 reduces to the MaskFormer paradigm. For the final model, we scale the training batch size from 2 to 128.}
\label{tab:ablation}
\vspace{-1em}
\end{table}

\paragraph{Multi-stage fusion and deep supervision drive substantial gains}
To quantify the impact of our architectural components, we performed ablation studies on our validation set, keeping training data and text encoder fixed across methods (Tab.~\ref{tab:ablation}). Transitioning from MaskFormer-inspired single-stage late fusion baselines (SAT + Qwen3-Embedding-4B) to earlier and repeated feature fusion across multiple decoder layers enables image features to interact with text prompts at finer stages of representation. Single-stage fusion paradigms (Mask2Former, MaskFormer) achieve up to 55.1 Dice, underscoring the limitations of late-only prompt integration. Incorporating VoxTell's multi-stage fusion increases Dice to 60.2–61.5, demonstrating that repeated cross-modal interactions across decoder stages substantially enhance prompt responsiveness. Adding deep supervision further boosts performance to 62.6 Dice, confirming that guiding intermediate representations strengthens alignment between visual and textual features. Finally, scaling the training batch size from 2 to 128 achieves the highest Dice, indicating that architecture and training scale jointly affect performance.

\begin{figure}[t]
  \centering
  \includegraphics[width=\linewidth]{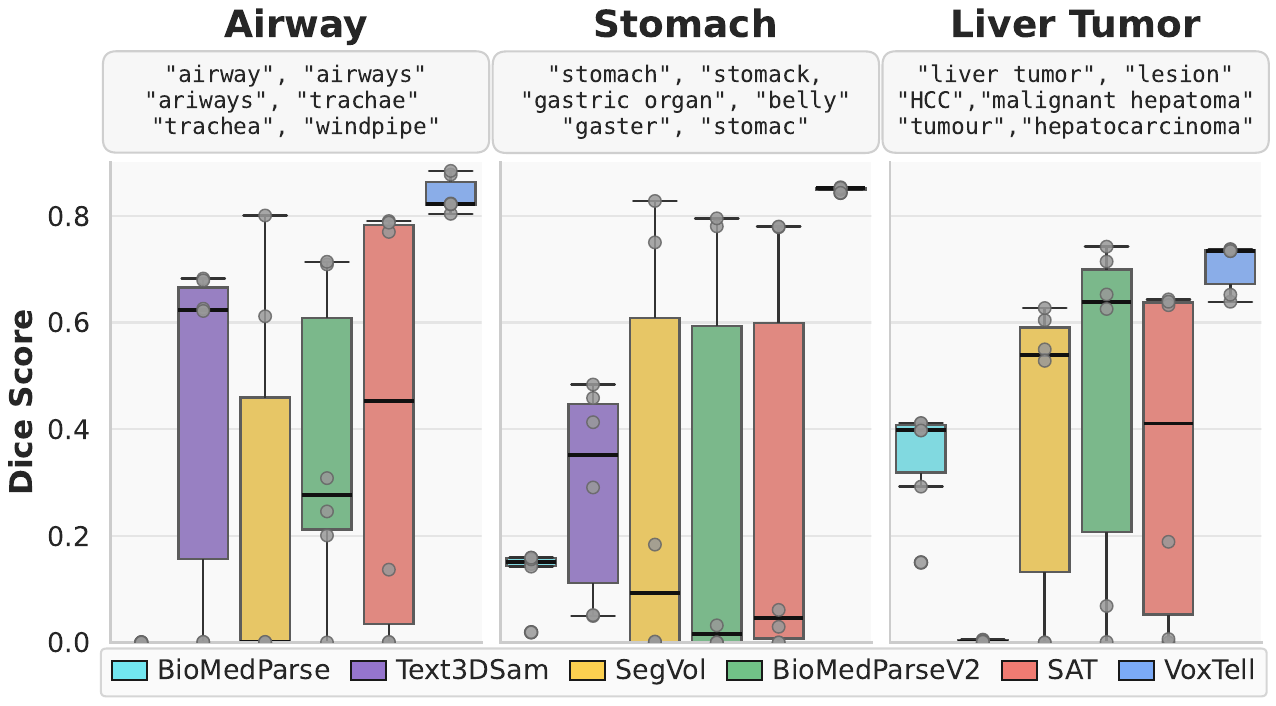}
  \caption{\textbf{Prompt Stability.} Dice score distributions of all methods across multiple textual prompts for the same anatomical structure. Competing methods exhibit high variability, often failing on certain synonyms or misspellings, while VoxTell maintains consistently high performance, even on prompts not seen during training.}
  \label{fig:prompt_stability}
  \vspace{-1em}
\end{figure}

\paragraph{Robustness to diverse prompts} We evaluate VoxTell’s sensitivity to variations in textual inputs, including synonyms, rephrasings, and minor spelling errors (Figure~\ref{fig:prompt_stability}). Although prior methods include prompt variation schemes during training, they often fail to generalize beyond a limited set of predefined label names, leading to poor performance even on simple alternative descriptions of common structures. In contrast, VoxTell maintains stable performance across all prompt formulations, despite many of them not appearing during training. This robustness stems from our comprehensive vocabulary enrichment and harmonization strategy, which integrates synonymous medical terms and related textual expressions, coupled with a strong pre-trained text encoder that maps diverse linguistic expressions to consistent embeddings. To further investigate the latter component, we perform complementary ablations (Appendix~\ref{sec:appendix_text_encoder_selection}) using state-of-the-art embedding models from the MTEB~\cite{muennighoff2022mteb} benchmark, confirming that embedding quality substantially impacts performance. VoxTell’s mid-sized encoder achieves an effective balance between robustness and computational efficiency, enabling consistent behavior across diverse natural-language queries.

\paragraph{Generalization to novel modalities and unseen concepts} We evaluate VoxTell on two challenging scenarios: (i) known concepts presented in a new imaging modality (cross-modality transfer), and (ii) entirely out-of-distribution concepts not seen during training (Tab.~\ref{tab:ood}). These tasks are particularly suited for text-promptable models, as classical segmentation networks cannot address cases without predefined classes. VoxTell reliably locates the object of interest and achieves strong Dice scores for unseen lesion types and across modality shifts, representing a significant step toward open-set generalization. While Dice scores on completely unseen concepts vary across structures (\eg 69.1 for esophageal cancer,  25.8 for bladder cancer), VoxTell consistently surpasses all prior approaches and  produces meaningful segmentations even in challenging cases, reflecting effective interpolation within its learned latent space. Qualitative examples in Fig.~\ref{fig:qualitative} highlight segmentations from familiar categories to new classes, including clinical language prompts, illustrating the model’s potential for open-world generalization.

\begin{figure}[t]
  \centering
  \setlength{\abovecaptionskip}{2pt} % adjust this value as needed
  \includegraphics[width=\linewidth]{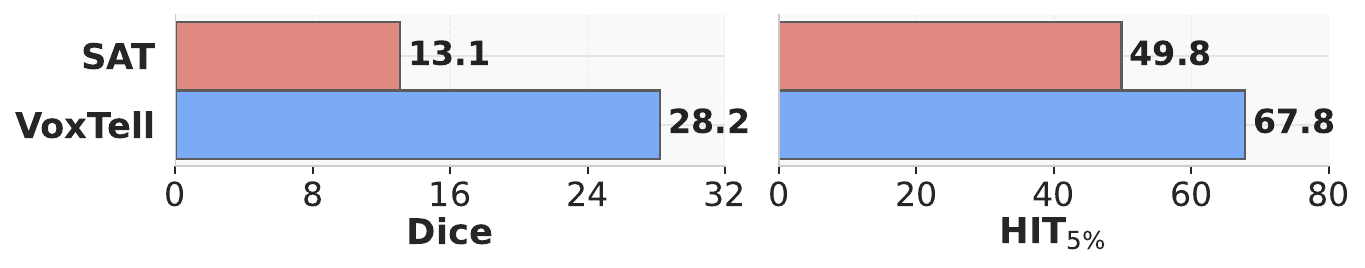}
  \caption{\textbf{Free-Text Segmentation on ReXGroundingCT.} Evaluation on the ReXGroundingCT benchmark~\cite{baharoon2025state} (validation set), which links radiology report findings to 3D segmentations in CT-RATE~\cite{ct-rate} chest CTs, assessing instance-level localization and segmentation from text. Following the benchmark protocol, both the current SoTA, SAT and VoxTell were fine-tuned on the training set. VoxTell outperforms SAT in Dice and hit-rate HIT$_{5\%}$ (the fraction of instances with Dice $\ge$ 5\%).}
  \label{fig:rex_grounding}
\end{figure}

\begin{table}[t]
\centering
\setlength{\tabcolsep}{2pt}
\begin{adjustbox}{max width=\columnwidth}
\begin{tabular}{l|ccc||cc|}
\toprule
 & \multicolumn{3}{c||}{Cross-Modality} & \multicolumn{2}{c}{Unknown Concept}\\
 \midrule
Structure & \shortstack{Breast \\ Cancer} & \shortstack{Pancreas\\ Tumor} & \shortstack{Sarcoma} & \shortstack{Bladder\\ Cancer} & \shortstack{Esophag. \\ Tumor} \\
Modality & \PET & \MRI & \MRI & \MRI & \CT  \\
\midrule
BioMedParse & \colorrange{0.75} & \colorrangeq{1.39} & \colorranger{36.20} & \colorranges{10.37} & \colorranged{17.86}  \\
Text3DSam & \colorrange{0.00} & \colorrangeq{4.52} & \colorranger{12.83} & \colorranges{2.04} & \colorranged{7.28} \\
SegVol & \colorrange{0.05} & \colorrangeq{14.64} & \colorranger{0.83} & \colorranges{0.08} & \colorranged{32.24} \\
BioMedParseV2 & \colorrange{0.00} & \colorrangeq{18.24} & \colorranger{7.68} & \colorranges{2.69} & \colorranged{16.56} \\
SAT & \colorrange{58.26} & \colorrangeq{19.25} & \colorranger{10.64} & \colorranges{9.56} & \colorranged{0.00} \\
\midrule
\textbf{VoxTell (ours)} & \textbf{\colorrange{72.27}} & \textbf{\colorrangeq{35.66}} & \textbf{\colorranger{40.34}} & \textbf{\colorranges{25.76}} & \textbf{\colorranged{69.07}}\\
\bottomrule
\end{tabular}
\end{adjustbox}
\caption{\textbf{Cross-Modality and unseen Concept Generalization.} VoxTell maintains strong Dice performance when transferring known structures to new modalities and on unseen related concepts, outperforming all prior methods. While performance varies across structures, the results demonstrate clear progress toward open-set generalization.}
\label{tab:ood}
\vspace{-1em}
\end{table}     

\begin{figure*}[t]
  \centering
  \includegraphics[width=\textwidth]{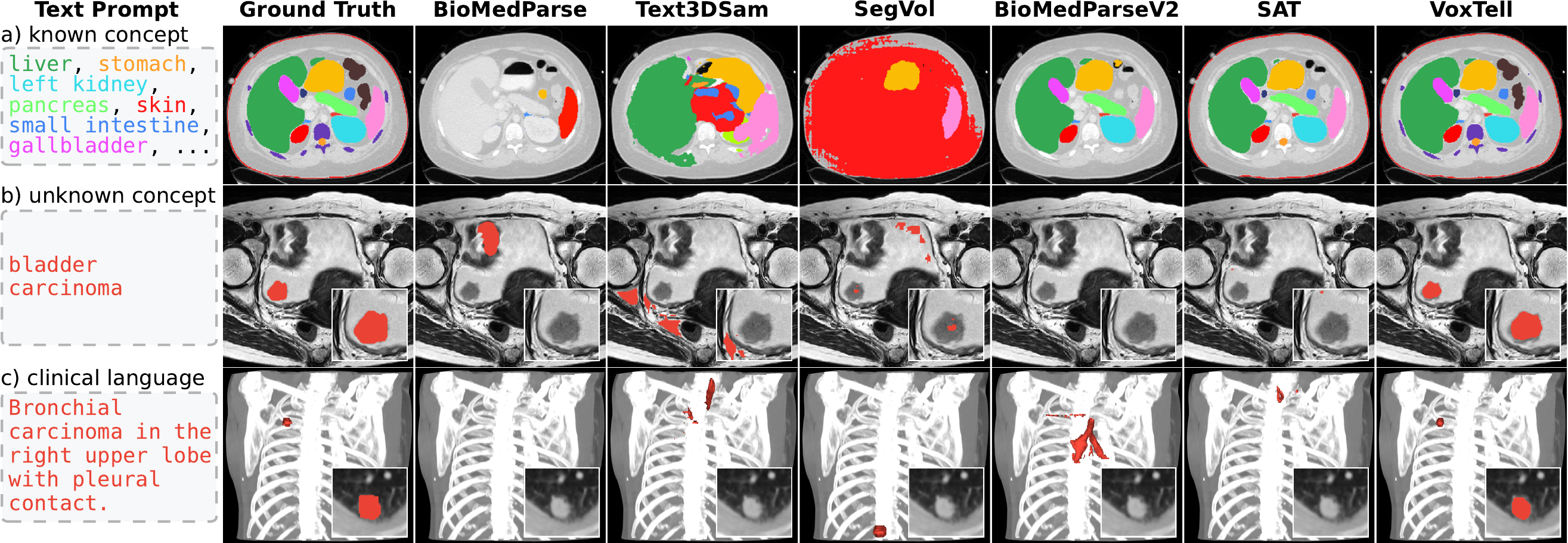}
\caption{\textbf{Qualitative comparison of text-prompted segmentation across varying prompt complexity.} (a) Known anatomical concepts, (b) unseen pathological structures, and (c) sentence-level clinical descriptions from in-house radiology reports. VoxTell produces accurate segmentations across all prompt types, while competing methods struggle on in-distribution prompts and fail on unseen or complex queries.}
  \label{fig:qualitative}
  \vspace{-1em}
\end{figure*}

\paragraph{Segmentation from real-world clinical reports} On our held-out radiotherapy cohort with sentence-level prompts extracted directly from radiology reports (203 patients, Appendix~\ref{sec:in_house_data}), VoxTell achieves 50.2 Dice, substantially outperforming SAT (0.0), BioMedParseV2 (1.2), and SegVol (8.1), despite all models having seen lung tumors during training (Fig.~\ref{fig:teaser}, right). The prompts include complex spatial and semantic descriptions, \eg "spiculated carcinoma in the left upper lobe with pleural contact." We attribute VoxTell's strong performance to the combination of training on diverse anatomical and pathological concepts, multi-stage vision–language fusion (Tab.~\ref{tab:ablation}) and robust text encoding (Appendix~\ref{sec:appendix_text_encoder_selection}), demonstrating successful interpretation of complex clinical language.

\paragraph{Leading model on the ReXGroundingCT benchmark} We further evaluate VoxTell on the public ReXGroundingCT benchmark~\cite{baharoon2025state} (validation split), which contains unique medical phrases describing spatially-specific findings in chest CT scans from CT-RATE~\cite{ct-rate}. Following the benchmark protocol, we fine-tune both VoxTell and the current top-performing SAT model on the ReXGroundingCT training set, augmented with our curated instance-specific dataset (Appendix~\ref{sec:appendix_instance_dataset_creation}). VoxTell achieves 28.2 Dice and 67.8 HIT$_{5\%}$, outperforming SAT's 13.1 Dice and 49.8 HIT$_{5\%}$, shown in Fig.~\ref{fig:rex_grounding}. Since both models were fine-tuned on the same data, the improvement likely stems from architectural differences, particularly VoxTell's multi-stage fusion design compared to SAT's single-stage approach. Integrating prompt information throughout the decoder produces spatially grounded, location-aware representations, allowing to differentiate anatomically similar structures based on spatial context (\eg left vs. right lung lesions). VoxTell establishes a new state-of-the-art for instance-specific segmentation from realistic, report-derived prompts.

\paragraph{Limitations} While VoxTell represents a significant step toward broader generalization across imaging modalities and semantically related unseen concepts, performance on fully out-of-distribution (OOD) cases remains challenging. For example, results on the Stanford Knee dataset~\cite{stanford_knee} (Appendix Tab.~\ref{tab:limitations}), which includes anatomical structures entirely absent from training, show that the model struggles with these concepts. In particular, the knee region is sparsely represented in the training set and features only a few segmentation targets, highlighting that the model’s text-guidance cannot extrapolate to entirely unfamiliar spatial or visual patterns. A similar limitation has been observed in 2D open‑vocabulary segmentation, where models fail to correctly identify or segment truly unseen classes and perform best on concepts closely related to known categories~\cite{openvocabsegmentation}. Future work could explore few-shot adaptation with minimal labeled examples, and leveraging text supervision from radiology report-image pairs, offering more diverse concepts than manual segmentation datasets.
\section{Conclusion}
\label{sec:conclusion}

We introduced VoxTell, a 3D vision–language segmentation model that enables robust, text-prompted volumetric segmentation from free-form clinical prompts. Through multi-stage vision–language fusion integrated across the decoder hierarchy, VoxTell achieves continuous alignment between textual queries and volumetric features. Trained on over 62K volumes across CT, PET, and MRI modalities encompassing over 1,000 structures, VoxTell achieves state-of-the-art performance on anatomical and pathological segmentation, outperforming both text-promptable methods and classical approaches like TotalSegmentator.\\

\noindent Beyond standard segmentation, VoxTell exhibits generalization capabilities across three dimensions: (i) robustness to diverse textual formulations, including synonyms, rephrasings, and spelling variations; (ii) cross-modality transfer, successfully segmenting known structures in unseen imaging modalities; and (iii) semantic extrapolation to related but unseen concepts, a significant step toward open-set 3D segmentation. Notably, VoxTell establishes new state-of-the-art results on segmentation from real-world radiology reports, demonstrating its potential for clinically relevant, text-driven medical image analysis.\\

\noindent While challenges remain in generalizing to entirely out-of-distribution anatomical regions, VoxTell represents meaningful progress toward flexible, language-driven 3D medical image segmentation. By bridging the gap between natural language and volumetric medical imaging, this work opens new avenues for intuitive clinical workflows and broader accessibility of advanced segmentation capabilities.

\section*{Acknowledgments}

M.R. is funded through a Google PhD Fellowship. The present contribution is supported by the Helmholtz Association under the joint research school "HIDSS4Health – Helmholtz Information and Data Science School for Health". This work was partly funded by Helmholtz Imaging (HI), a platform of the Helmholtz Incubator on Information and Data Science. This work was partially supported by RACOON, funded by “NUM 2.0” (FKZ: 01KX2121) as part of the Racoon Project. This work was supported by the Helmholtz Association's Initiative and Networking Fund on the HAICORE@FZJ partition.

\newpage
{
    \small
    \bibliographystyle{ieeenat_fullname}
    \bibliography{main}
}
\clearpage
\renewcommand{\thefootnote}{\arabic{footnote}}
\setcounter{page}{1}
\maketitlesupplementary
\appendix

\MarkAppendixInTOC  
\setcounter{tocdepth}{2}
\AppendixOnlyTOC 

\section{Implementation Details}
\label{sec:appendix_implementation_details}

\subsection{Training Configuration}
All experiments were trained under a consistent configuration to ensure reproducibility and fair comparisons across ablations and baselines. For ablation studies, a single GPU with a batch size of 2 was used, while the final VoxTell model was trained on 64 NVIDIA A100 GPUs (40GB each) with a batch size of 128, achieving 64$\times$ higher throughput. We leverage the nnU-Net~\cite{isensee_nnu-net_2021} framework, which provides a robust 3D segmentation training pipeline. Each model was trained for 2,000 epochs, following the convention of 250 iterations per epoch, using stochastic gradient descent (SGD) with an initial learning rate of $1\times10^{-4}$, decayed according to a polynomial schedule. Input volumes were processed as $192^3$ voxel patches. Standard nnU-Net data augmentation strategies were applied, with the exception of mirroring along the left–right axis to avoid ambiguities in laterality-sensitive anatomical structures.

\subsection{Model Architecture}
The vision backbone is a ResEncL encoder~\cite{nnunet_revisited} comprising six layers, producing hierarchical feature maps with channel dimensions \verb+[32, 64, 128, 256, 320, 320]+. The prompt decoder is a standard transformer using cross attention with six layers, eight heads and a query dimension of 2048. Cross-modal integration is performed via a two-layer MLP vision–text adapter with hidden dimension 2048, which aligns the dimensions of textual features produced by the prompt decoder with visual features. The text encoder is a frozen Qwen3-Embedding-4B~\cite{qwen3embedding} model, converting natural language prompts into fixed embeddings. During training, all text embeddings are precomputed for efficiency. We use the following instruction prompt:\\

\noindent \textit{``Instruct: Given an anatomical term query, retrieve the precise anatomical entity and location it represents. Query: [text input].''}\\

\noindent Initial experimentation revealed that this prompt ensures consistent mapping of anatomical and pathological terms to similar embeddings.

\subsection{Training Strategy}
Each training image is queried with three textual prompts as this is the maximum that fits on an NVIDIA A100 GPU (40GB) during training with batch size 2 per GPU. We use two positive prompts corresponding to structures present in the volume, and one negative prompt corresponding to an absent structure. The negative prompt is critical for teaching the model to output empty masks when the target is not present. Foreground structures are oversampled at an 85\% probability to increase the frequency of patches containing segmentation targets. For positive text prompts, we randomly sample synonyms and rephrasings for each concept as prompts, while emphasizing the main term (\eg “liver” over “hepatic organ”), by selecting the default name with 25\% probability and a rephrased variant with 75\%. The segmentation objective combines Dice loss with binary cross-entropy (BCE) loss to optimize both volumetric and pixel-wise accuracy. Deep supervision is applied at all five decoder scales, using nnU-Net’s default weights $\lambda_s = [1, 1/2, 1/4, 1/8, 1/16]$. Vision–language feature fusion is applied at all scales, ensuring repeated cross-modal interactions throughout the decoder and enhancing alignment between textual prompts and visual features.

\section{Dataset Details}
\label{sec:dataset_details}

\begin{figure*}[t]
  \centering
  \makebox[\textwidth][c]{%
    \includegraphics[width=1.1\textwidth]{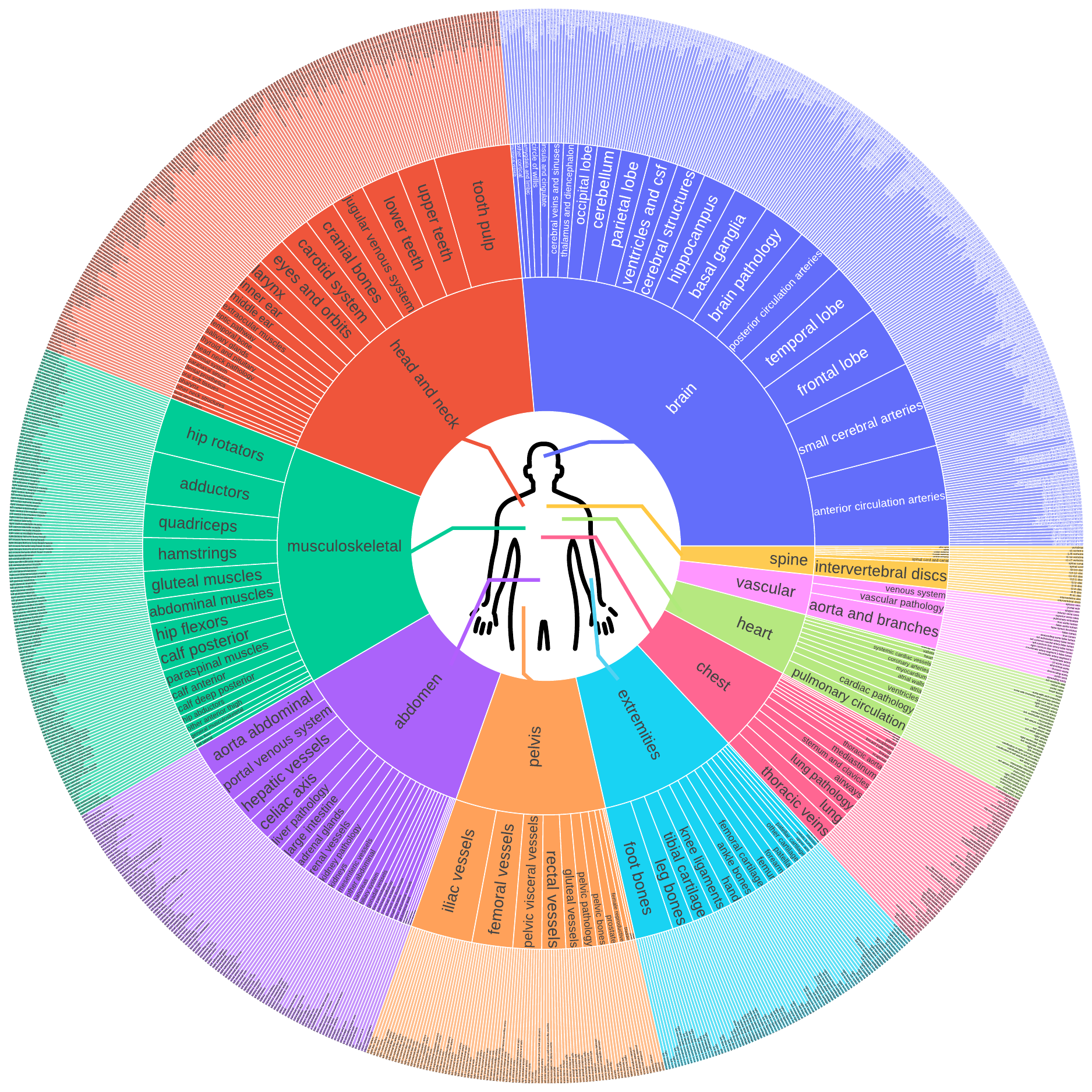}%
  }
  \caption{\textbf{Comprehensive overview of the anatomical and pathological concepts} included in our multi-modality 3D medical imaging dataset. The dataset aggregates 158 public sources, covering CT, MRI, and PET scans of over 62,000 volumetric images across brain, head and neck, thorax, abdomen, pelvis, musculoskeletal, extremities and vascular systems. The visualization highlights both coarse and fine-grained structures, ranging from major organs to substructures and lesions, demonstrating the semantic diversity leveraged for language-conditioned 3D segmentation in VoxTell.}
  \label{fig:dataset_circle}
\end{figure*}
% ========== TABLE 1 ==========
\begin{table*}[h]
\centering
\scriptsize
\resizebox{\textwidth}{!}{
\begin{tabular}{lcccrr}
\toprule
Name & Images & Categories & Modality & Target & Link \\
\midrule
Decatlon Task 2 \cite{simpson2019large} & 20 & 1 & MRI & Left Atrium & \url{http://medicaldecathlon.com} \\
Decatlon Task 3 \cite{simpson2019large} & 131 & 2 & CT & Liver, L. Tumor & \url{http://medicaldecathlon.com} \\
Decatlon Task 4 \cite{simpson2019large} & 260 & 2 & MRI & Hippocampus & \url{http://medicaldecathlon.com} \\
Decatlon Task 5 \cite{simpson2019large} & 32 & 3 & MRI & Prostate & \url{http://medicaldecathlon.com} \\
Decatlon Task 6 \cite{simpson2019large} & 63 & 1 & CT & Lung Cancer & \url{http://medicaldecathlon.com} \\
Decatlon Task 7 \cite{simpson2019large} & 281 & 2 & CT & Pancreas, P. Tumor & \url{http://medicaldecathlon.com} \\
Decatlon Task 8 \cite{simpson2019large} & 303 & 2 & CT & Hepatic Vessel, H. Tumor & \url{http://medicaldecathlon.com} \\
Decatlon Task 9 \cite{simpson2019large} & 41 & 1 & CT & Spleen & \url{http://medicaldecathlon.com} \\
Decatlon Task 10 \cite{simpson2019large} & 126 & 1 & CT & Colon Cancer & \url{http://medicaldecathlon.com} \\
ISLES2015 \cite{maier2015extra} & 28 & 1 & MRI & Stroke Lesions & \url{http://www.isles-challenge.org/ISLES2015} \\
BTCV \cite{Landman2015MALBCV} & 30 & 13 & CT & 13 Abdominal Organs & \url{https://www.synapse.org/Synapse:syn3193805/wiki/217789} \\
RibSeg \cite{ribfracchallenge2025,ribfracclinical2020} & 654 & 25 & CT & Ribs & \url{https://ribfrac.grand-challenge.org} \\
AortaSeg24 \cite{IMRAN2024102470} & 50 & 23 & CT & Aorta & \url{https://aortaseg24.grand-challenge.org} \\
LIDC \cite{lidc} & 1010 & 1 & CT & Lung Lesion & \url{https://www.cancerimagingarchive.net/collection/lidc-idri} \\
CTPelvic1k \cite{Liu2020PelvicSegmentation} & 1106 & 3 & CT & Pelic Bones & \url{https://github.com/MIRACLE-Center/CTPelvic1K} \\
Promise \cite{litjens2014evaluation} & 50 & 1 & MRI & Prostate & \url{https://zenodo.org/records/8026660} \\
Duke Liver \cite{luo2024automatic} & 310 & 1 & MRI & Liver & \url{https://zenodo.org/records/7774566} \\
ACDC \cite{bernard2018deep} & 200 & 3 & MRI & Cardiac Structures & \url{https://www.creatis.insa-lyon.fr/Challenge/acdc} \\
AbdOrgSegm \cite{gibson2018automatic,gibson2018automatic,roth2016data,roth2015deeporgan,clark2013cancer,xu2016evaluation} & 63 & 8 & CT & Abdominal Organs & \url{https://zenodo.org/records/1169361} \\
CHAOS \cite{CHAOS2021,kavur2019} & 60 & 4 & MR & liver, right kidney, left kidney, spleen & \url{https://chaos.grand-challenge.org/Data} \\
OpenMind Tissue \cite{wald2025openmind,billot_synthseg_2023} & 3644 & 32 & MRI, CT & Brain Structures & \url{https://github.com/BBillot/SynthSeg} \\
StructSeg Task 1 \cite{hongsheng_li_2020_3718885} & 50 & 22 & CT & 23 Head and Neck Structures & \url{https://structseg2019.grand-challenge.org} \\
StructSeg Task 2 \cite{hongsheng_li_2020_3718885} & 50 & 1 & CT & Nasopharynx Cancer & \url{https://structseg2019.grand-challenge.org} \\
StructSeg Task 3 \cite{hongsheng_li_2020_3718885} & 50 & 6 & CT & 6 Thoracic Structures & \url{https://structseg2019.grand-challenge.org} \\
StructSeg Task 4 \cite{hongsheng_li_2020_3718885} & 50 & 1 & CT & Lung Tumor & \url{https://structseg2019.grand-challenge.org} \\
COVID-19-20 \cite{An2020CTImages,roth2022rapid} & 199 & 1 & CT & Covid & \url{https://covid-segmentation.grand-challenge.org/COVID-19-20} \\
SegTHOR \cite{lambert2020segthor} & 40 & 4 & CT & Heart, Aorta, Trachea, Esophagus & \url{https://competitions.codalab.org/competitions/21145} \\
FETA \cite{payette2021automatic} & 40 & 7 & MRI & Brain Regions & \url{https://fetachallenge.github.io/pages/Data_description} \\
ISLES2022 \cite{hernandez2022isles} & 250 & 1 & MRI & Stroke Lesion & \url{https://zenodo.org/records/7153326} \\
LGGMRISeg \cite{buda2019association} & 110 & 1 & MRI & Brain Tumor & \url{https://www.kaggle.com/datasets/mateuszbuda/lgg-mri-segmentation/data} \\
NIH-Pan \cite{roth2015deeporgan} & 82 & 1 & CT & Pancreas & \url{https://wiki.cancerimagingarchive.net/display/Public/Pancreas-CT} \\
M-CRIB \cite{alexander2019desikan} & 10 & 100 & MRI & Neonatal Brain Atlas & \url{https://osf.io/4vthr} \\
CAP \cite{kadish2009rationale} & 1637 & 1 & MRI & Left Ventricle & \url{https://www.cardiacatlas.org/lv-segmentation-challenge} \\
AtriaSeg2018 \cite{xiong2021global} & 154 & 2 & MRI & Left Atrium & \url{https://www.cardiacatlas.org/atriaseg2018-challenge/atria-seg-data} \\
Spine-Mets \cite{pieper2024spinemets} & 55 & 17 & CT & Vertebrae & \url{https://www.cancerimagingarchive.net/collection/spine-mets-ct-seg} \\
VerSe2019 \cite{loffler2020vertebral,sekuboyina2020labeling,Sekuboyina_2021} & 80 & 26 & CT & Vertebrae & \url{https://osf.io/jtfa5} \\
VerSe2020 \cite{loffler2020vertebral,sekuboyina2020labeling,Sekuboyina_2021} & 61 & 25 & CT & Vertebrae & \url{https://verse2020.grand-challenge.org} \\
WMHSegChallenge \cite{kuijf2019standardized} & 60 & 1 & MRI & White Matter Hyperintensities & \url{https://dataverse.nl/dataset.xhtml?persistentId=doi:10.34894/AECRSD} \\
NCI-ISBI-Prostate \cite{bloch2015nci} & 60 & 3 & MRI & Prostate Zones & \url{http://doi.org/10.7937/K9/TCIA.2015.zF0vlOPv} \\
OASIS \cite{marcus2007open} & 436 & 3 & MRI & Brain Regions & \url{https://sites.wustl.edu/oasisbrains/home/oasis-1} \\
BraTS24 Task 1 Glioma \cite{deverdier20242024braintumorsegmentation} & 1350 & 4 & MRI & Glioblastoma & \url{https://www.synapse.org/Synapse:syn53708249/wiki/627500} \\
BraTS2024 Task 2 Africa \cite{adewole2023braintumorsegmentationbrats} & 60 & 3 & MRI & Brain Tumor & \url{https://www.synapse.org/Synapse:syn53708249/wiki/627501} \\
BraTS24 Task 3 Meningioma \cite{labella2025analysis2024bratsmeningioma} & 500 & 1 & MRI & Brain Tumor & \url{https://www.synapse.org/Synapse:syn53708249/wiki/627503} \\
BraTS24 Task 4 Brain Metastases \cite{moawad2023brain} & 652 & 3 & MRI & Brain Metastases & \url{https://www.synapse.org/Synapse:syn53708249/wiki/627504} \\
BraTS24 Task 5 Pediatric \cite{kazerooni2024braintumorsegmentationpediatrics} & 261 & 4 & MRI & Brain Tumor & \url{https://www.synapse.org/Synapse:syn53708249/wiki/627505} \\
LNQ2023 \cite{dorent2024lnq} & 393 & 1 & CT & Mediastinal Lymph Nodes & \url{https://lnq2023.grand-challenge.org/data} \\
MediaLymph \cite{bouget2021mediastinal} & 15 & 1 & CT & Mediastinal Lymph Nodes & \url{https://github.com/dbouget/ct_mediastinal_structures_segmentation} \\
MediaStruct \cite{bouget2019semantic} & 15 & 12 & CT & Mediastinal Structures & \url{https://github.com/dbouget/ct_mediastinal_structures_segmentation} \\
CT Lymph Nodes \cite{roth2014new} & 176 & 1 & CT & Mediastinal Lymph Nodes & \url{https://www.cancerimagingarchive.net/collection/ct-lymph-nodes} \\
MAMA MIA \cite{mama_mia} & 1506 & 1 & MRI & Breast Tumor & \url{https://www.synapse.org/Synapse:syn60868042/wiki/628716} \\
ATM2022 \cite{atm22} & 0 & 1 & CT & Airway Tree & \url{https://atm22.grand-challenge.org} \\
Atlas Bourgogne \cite{quinton2023tumour} & 60 & 2 & MRI & Liver, Tumor & \url{https://atlas-challenge.u-bourgogne.fr} \\
Couinaud \cite{zhang2023anatomical} & 193 & 8 & CT & Liver Segments & \url{https://www.kaggle.com/datasets/louisgv/couinaud-liver-segmentation} \\
CURVAS \cite{RieraMarin2024} & 60 & 3 & CT & Pancreas, Kidney, Liver & \url{https://curvas.grand-challenge.org/curvas-dataset} \\
MMs \cite{campello2021multi} & 300 & 3 & MRI & Cardiac Structures & \url{https://www.ub.edu/mnms} \\
Emidec \cite{lalande2020emidec} & 100 & 4 & MRI & Cardiac Structures & \url{https://emidec.com} \\
Kipa22 \cite{he2021meta} & 70 & 4 & CT & Kidney, Vessel, Tumor & \url{https://kipa22.grand-challenge.org} \\
MrBrains18 \cite{kuijf2024mr} & 30 & 9 & MRI & Brain Structures & \url{https://mrbrains18.isi.uu.nl/index.html} \\
OrCaScore \cite{wolterink2016evaluation} & 32 & 3 & CT & Heart Calcifications & \url{https://orcascore.grand-challenge.org} \\
TriALS Task 2 \cite{elbatel_2024_10992127} & 240 & 2 & CT & Liver, L. Tumor & \url{https://www.synapse.org/Synapse:syn53285416} \\
Parse22 \cite{luo2023efficient} & 100 & 1 & CT & Pulmonary Artery & \url{https://parse2022.grand-challenge.org/Parse2022} \\
PDDCA \cite{raudaschl2017evaluation} & 47 & 9 & CT & Head and Neck Structures & \url{https://www.imagenglab.com/newsite/pddca} \\
Aortic Dissection \cite{Mayer2024} & 39 & 2 & CT & True and False Aortic Lumen & \url{https://figshare.com/ndownloader/articles/22269091/versions/1} \\
ProstateEdgeCases \cite{kanwar2023stress} & 131 & 5 & CT & Bladder, Prostate, Rectum & \url{https://doi.org/10.7937/QSTF-ST65} \\
SKI10 \cite{ski10} & 100 & 4 & MRI & Cartilage, Bone & \url{https://ski10.grand-challenge.org} \\
Spider \cite{spider} & 447 & 19 & MRI & Lumbar Spine & \url{https://zenodo.org/records/10159290} \\
VALDO Task 2 \cite{sudre2024valdo} & 72 & 1 & MRI & Cerebral Microbleed & \url{https://valdo.grand-challenge.org/Task2} \\
BONBID-HIE \cite{rina_bao_2023_10602767} & 85 & 1 & MRI & Brain Lesion & \url{https://zenodo.org/records/1060276} \\
UPENN-GBM \cite{upenn-gbm} & 147 & 3 & MRI & Brain Edema and Tumor & \url{https://www.cancerimagingarchive.net/collection/upenn-gbm} \\
LNDb \cite{lndb} & 236 & 1 & CT & Lung Nodule & \url{https://lndb.grand-challenge.org} \\
ReMIND \cite{remind} & 213 & 1 & MRI & Brain Resection & \url{https://www.cancerimagingarchive.net/collection/remind} \\
ProstateX \cite{litjens2017prostatex} & 140 & 2 & MRI & Prostate Tumor & \url{https://www.cancerimagingarchive.net/collection/prostatex} \\
Prostate158 \cite{ADAMS2022105817} & 188 & 1 & MRI & Prostate, P. Tumor & \url{https://zenodo.org/records/6481141} \\
TotalSegmentatorMRI \cite{wasserthal_2025_14710732} & 616 & 50 & MRI & Whole Body Structures & \url{https://zenodo.org/records/14710732} \\
RiderLungCT \cite{Zhao2015_RiderLungCT} & 65 & 1 & CT & Lung Cancer & \url{https://www.cancerimagingarchive.net/collection/rider-lung-ct} \\
LiverMets \cite{colorectal-liver-metastases} & 171 & 4 & CT & Liver Metastases & \url{https://doi.org/10.7937/QXK2-QG03} \\
HippocampusSubfield \cite{Li2024} & 20 & 14 & MRI & Hippocampus Structures & \url{https://plus.figshare.com/ndownloader/articles/26075713/versions/1} \\
PANORAMA \cite{alves2024panorama} & 482 & 6 & CT & Pancreas Structures & \url{https://panorama.grand-challenge.org} \\
TOM500 \cite{Song2025} & 400 & 9 & MRI & Eye Structures & \url{https://springernature.figshare.com/ndownloader/files/49499655} \\
\bottomrule
\end{tabular}
}
\caption{\textbf{Overview of training datasets part 1.} Overview of the 158 datasets used for model training (Part 1/2), covering names, image counts, number of categories, modalities, targets, and access links.}
\vspace*{12em}
\label{tab:appendix_train_datasets_1}
\end{table*}

% ========== TABLE 2 ==========
\begin{table*}[h]
\centering
\scriptsize
\resizebox{\textwidth}{!}{
\begin{tabular}{lcccrr}
\toprule
Name & Images & Categories & Modality & Target & Link \\
\midrule
WAW-TACE \cite{bartnik_2024_12741586} & 114 & 1 & CT & Hepatocellular Carcinoma & \url{https://zenodo.org/records/12741586} \\
LiverHccSeg \cite{gross_2023_8179129} & 34 & 2 & MRI & Hepatocellular carcinoma & \url{https://zenodo.org/records/8179129} \\
Enhance-PET \cite{enhancepet1_ferrara2025sharing,enhancepet2_shiyamsundar2022fully} & 1597 & 132 & CT, PET & Whole Body Organs & \url{https://github.com/ENHANCE-PET/MOOSE} \\
LLD-MMRI \cite{LLD-MMRI,MedSAM2} & 3984 & 1 & MRI & Liver Lesion & \url{https://huggingface.co/datasets/wanglab/LLD-MMRI-MedSAM2} \\
AIIB23 \cite{li2022human,nan2023fuzzy} & 120 & 1 & CT & Airway & \url{https://codalab.lisn.upsaclay.fr/competitions/13238} \\
BTCV-Cervix \cite{landman2015miccai} & 30 & 4 & CT & Abdominal Organs & \url{https://www.synapse.org/Synapse:syn3193805/wiki/217752} \\
CARE25-MyoPS \cite{zhuang2019multivariate,qiu2023myops,ding2023aligning} & 449 & 5 & MRI & Cardiac Structures, C. Edema and Scars & \url{https://zmic.org.cn/care_2025/track2} \\
CARE25-WHS \cite{Zhuang2016MSMMA,Zhuang2019MvMM,GAO2023BayeSeg} & 86 & 7 & CT, MRI & Cardiac Structures & \url{https://zmic.org.cn/care_2025/track3} \\
cSeg2022 \cite{sun2023self} & 13 & 3 & MRI & Brain Regions & \url{https://tarheels.live/cseg2022} \\
ImageCAS \cite{zeng2023imagecas} & 1000 & 1 & CT & Coronary Artery & \url{https://www.kaggle.com/datasets/xiaoweixumedicalai/imagecas/data} \\
ToothFairy 3 \cite{ToothFairy3_2025CVPR,ToothFairy3_2024TMI,ToothFairy3_2024IEEEACCESS} & 532 & 77 & CT & Dental Structures & \url{https://ditto.ing.unimore.it/toothfairy3} \\
FUMPE \cite{masoudi_saadatmand-tarzjan_2018} & 35 & 1 & CT & Pulmonary Embolism & \url{https://www.kaggle.com/datasets/andrewmvd/pulmonary-embolism-in-ct-images} \\
SLAWT \cite{karim2018algorithms} & 10 & 1 & CT & Left Atrium, Atrium Wall & \url{https://www.doc.ic.ac.uk/~rkarim/la_lv_framework/wall/index.html} \\
LiverCirrMRI \cite{jha2025large} & 676 & 1 & MRI & Liver & \url{https://osf.io/cuk24} \\
3D-IRCADb-01 \cite{soler20103d} & 20 & 31 & CT & Abdominal Organs, Liver Tumor & \url{https://www.ircad.fr/research/data-sets/liver-segmentation-3d-ircadb-01} \\
ImageTBAD \cite{yao2021imagetbad} & 100 & 3 & CTA & Aorta True Lumen, False Lumen, Thrombus & \url{https://www.kaggle.com/datasets/xiaoweixumedicalai/imagetbad} \\
ImageCHD \cite{xu2020imagechd} & 110 & 7 & CTA & Heart, Vascular Structures & \url{https://www.kaggle.com/datasets/xiaoweixumedicalai/imagechd} \\
MS-Brain-MRI-Lesion \cite{ms_flair_muslim2022brain} & 60 & 1 & MRI & Multiple Sclerosis Lesion & \url{https://data.mendeley.com/datasets/8bctsm8jz7/1} \\
ImageALCAPA \cite{image_ALPACA} & 30 & 7 & CT & Cardiac Structures & \url{https://www.kaggle.com/datasets/xiaoweixumedicalai/imagealcapa} \\
ImageTAAD \cite{song2025domain} & 120 & 35 & CT & Abdominal Structures & \url{https://www.kaggle.com/datasets/xiaoweixumedicalai/imagetaad} \\
Crossmoda2022 \cite{shapey2021segmentation} & 210 & 2 & MRI & vestibular Schwannoma, Cochlea & \url{https://zenodo.org/records/6504722} \\
Atlas22 \cite{liew2022large} & 655 & 1 & MRI & lesion & \url{https://bids.neuroimaging.io} \\
PulmonaryVessels \cite{10.1007/978-3-031-78198-8_28} & 106 & 2 & CT & Pulmonary vessels & \url{https://www.kaggle.com/datasets/xiaoweixumedicalai/mytest} \\
HOCMvalvesSeg \cite{zheng2024automatic} & 27 & 7 & CT & Cardiac structures & \url{https://www.kaggle.com/datasets/xiaoweixumedicalai/hocmvalvesseg} \\
CTSpine1K \cite{deng2024ctspine1klargescaledatasetspinal} & 1005 & 25 & CT & Vertebrae & \url{https://huggingface.co/datasets/alexanderdann/CTSpine1K} \\
Han-Seg23-MR \cite{podobnik2024han} & 42 & 30 & MRI & Head Neck Structures & \url{https://zenodo.org/records/7442914\#.ZBwp7XbMJaR} \\
HNTSMRG24 \cite{wahid_2024_11199559} & 150 & 2 & MRI & Head Neck Lesions & \url{https://zenodo.org/records/11199559} \\
IVDM3Seg \cite{chen20143d} & 16 & 1 & MRI & Intervertebral Discs & \url{https://ivdm3seg.weebly.com} \\
KiTS23 \cite{heller2023kits21} & 489 & 4 & CT & Kidney, K. Tumor, K. Cyst & \url{https://github.com/neheller/kits23} \\
AutoPet2 \cite{gatidis2022whole} & 1014 & 1 & CT, PET & Lesions & \url{https://autopet-ii.grand-challenge.org} \\
AutoPet3 \cite{gatidis2022fdgpet,ingrisch2024automated} & 597 & 1 & CT, PET & Lesions & \url{https://autopet-iii.grand-challenge.org} \\
AMOS \cite{ji2022amos} & 360 & 16 & CT & Abdominal Organs & \url{https://zenodo.org/record/7155725\#.Y0OOCOxBztM} \\
RHUH-GBM \cite{cepeda2023rhuhgbm} & 120 & 3 & MRI & Brain Tumor & \url{https://www.cancerimagingarchive.net/collection/rhuh-gbm} \\
AutoPet4 \cite{kuestner2025longitudinal} & 670 & 1 & CT & Lesions & \url{https://fdat.uni-tuebingen.de/records/qwsry-7t837} \\
CPTAC-HNSCC \cite{rozenfeld2023cptac} & 245 & 5 & CT & Head Neck Lesions & \url{https://doi.org/10.7937/PFEC-T641} \\
BrainTRGammaKnife \cite{wang2023braintrgammaknife,wang2023brain} & 76 & 1 & MRI & Brain Lesions & \url{https://www.cancerimagingarchive.net/collection/brain-tr-gammaknife} \\
LAScarQS24 Task 1 \cite{li2022medical,li2020atrial,li2022atrialjsqnet,li2021atrialgeneral} & 60 & 2 & MRI & Left Atrium, Atrial Scars & \url{https://zmic.org.cn/care_2024/track2} \\
LAScarQS24 Task 2 \cite{li2022medical,li2020atrial,li2022atrialjsqnet,li2021atrialgeneral} & 130 & 1 & MRI & Left Atrium & \url{https://zmic.org.cn/care_2024/track2} \\
QUADRA-HC \cite{gutschmayer2025wholebody} & 96 & 58 & CT & Whole Body Organs & \url{https://zenodo.org/records/16686025} \\
PancreasSegMRI T1 \cite{zhang2024large} & 385 & 1 & MRI & Pancreas & \url{https://osf.io/kysnj} \\
TopCoW24 \cite{topcowchallenge} & 250 & 13 & CT & Vessel Components of CoW & \url{https://topcow24.grand-challenge.org} \\
MRISegmentator & 540 & 45 & MRI & Whole Body Organs & \url{https://github.com/rsummers11/MRISegmenter} \\
TopBrain-CT \cite{topcowchallenge} & 25 & 40 & CT & Brain Structures & \url{https://topbrain2025.grand-challenge.org} \\
TopBrain-MR \cite{topcowchallenge} & 25 & 42 & MRI & Brain Structures & \url{https://topbrain2025.grand-challenge.org} \\
LISA25 \cite{lepore2025low} & 79 & 8 & MRI & Brain Structures & \url{https://www.synapse.org/Synapse:syn65670170/wiki/631438} \\
Lower Extremity Muscles \cite{andreassen2023three} & 78 & 126 & MRI & Musculoskeletal Structures & \url{https://digitalcommons.du.edu/visiblehuman} \\
CMRxMotion \cite{wang2022extremecardiacmrianalysis} & 138 & 3 & MRI & Cardiac Structures & \url{https://cmr.miccai.cloud} \\
BreastDivider \cite{rokuss2025divide} & 200 & 2 & MRI & Left Breast, Right Breast & \url{https://huggingface.co/datasets/Bubenpo/BreastDividerDataset} \\
TotalSegmentatorV2 \cite{wasserthal2023totalsegmentator} & 1228 & 120 & CT & Whole Body Structures & \url{https://zenodo.org/records/10047292} \\
Hecktor2022 \cite{10.1007/978-3-031-27420-6_1} & 524 & 2 & CT, PET & Head Neck Tumor & \url{https://hecktor.grand-challenge.org} \\
Instance2022 \cite{instance22} & 100 & 1 & CT & Intracranial Hemorrhage & \url{https://instance.grand-challenge.org} \\
MS Ljubljana \cite{lesjak2018novel,galimzianova2016stratified} & 264 & 1 & MRI & Muliple Sclerosis Lesion & \url{https://lit.fe.uni-lj.si/en/research/resources/3D-MR-MS/} \\
FLARE2022 \cite{FLARE22-LDH2024} & 50 & 13 & CT & Abdominal Organs & \url{https://flare22.grand-challenge.org} \\
SegRap23 Task 1 \cite{luo2023segrap2023} & 120 & 45 & CT & Head Neck Structures & \url{https://drive.google.com/drive/folders/115mzmNlZRIewnSR2QFDwW_-RkNM0LC9D} \\
SegA \cite{Radl2022} & 56 & 1 & CT & Aortic Vessel Tree & \url{https://figshare.com/ndownloader/articles/14806362/versions/1} \\
WORD \cite{luo2022word,liao2023comprehensive} & 120 & 17 & CT & Abdominal Organs & \url{https://github.com/HiLab-git/WORD} \\
AbdomenCT1K \cite{Ma-2021-AbdomenCT-1K} & 996 & 4 & CT & Liver, Kidney, Spleen, Pancreas & \url{https://zenodo.org/records/7860267} \\
DAP-ATLAS \cite{jaus2023towards} & 533 & 143 & CT & Abdominal Organs & \url{https://www.synapse.org/\#!Synapse:syn52287632.1/datasets} \\
TORG \cite{rister2020ct} & 140 & 4 & CT & Liver, Bladder, Lungs, Kidneys, Bone, Brain & \url{https://www.cancerimagingarchive.net/collection/ct-org} \\
HanSeg-CT \cite{HaNSeg_dataset,podobnik2024han} & 42 & 30 & CT & Head Neck Structures & \url{https://zenodo.org/records/7442914} \\
MU-Glioma-Post \cite{yaseen2025muglioma} & 593 & 4 & MRI & Brain Tumor & \url{https://www.cancerimagingarchive.net/collection/mu-glioma-post} \\
ACRIN-HN \cite{kinahan2019acrin6685} & 67 & 1 & CT, PET & Head Neck Lesion & \url{https://www.cancerimagingarchive.net/collection/acrin-hnscc-fdg-pet-ct} \\
Head-and-Neck-PET-CT \cite{vallieres2017headneck} & 34 & 1 & CT, PET & Head Neck Lesion & \url{https://www.cancerimagingarchive.net/collection/head-neck-pet-ct} \\
NSCLC-Radiogenomics \cite{bakr2017nsclc} & 69 & 1 & CT, PET & Lung Lesion & \url{https://www.cancerimagingarchive.net/collection/nsclc-radiogenomics} \\
Soft-Tissue-Sarcoma \cite{vallieres2015sarcoma} & 42 & 1 & CT, PET & Soft Tissue Sarcoma & \url{https://www.cancerimagingarchive.net/collection/soft-tissue-sarcoma} \\
TCGA-Lung \cite{albertina2016tcgaluad} & 5 & 1 & CT, PET & Lung Tumor & \url{https://www.cancerimagingarchive.net/collection/tcga-luad} \\
MSCMRSeg \cite{ZHUANG2022102528} & 25 & 3 & MRI & Cardiac Structures & \url{https://zmiclab.github.io/zxh/0/mscmrseg19/} \\
DeepLesion \cite{yan2018deeplesion} & 1093 & 1 & CT & Lesion & \url{https://nihcc.app.box.com/v/DeepLesion/} \\
Covid19CTLung \cite{ma2020covid19} & 10 & 1 & CT & Covid & \url{https://zenodo.org/records/3757476} \\
PANTHER Task 2 \cite{betancourt_tarifa_2025_15192302} & 50 & 2 & MRI & Pancreas, P. Tumor & \url{https://zenodo.org/records/15192302} \\
NSCLC-PleuralEffusion \cite{kiser2020plethora,aerts2019nsclc} & 78 & 1 & CT & Pleural Effusion & \url{https://www.cancerimagingarchive.net/analysis-result/plethora} \\
NSCLC-Radiomics \cite{aerts2014nsclc} & 78 & 1 & CT & Lung Tumor & \url{https://www.cancerimagingarchive.net/collection/nsclc-radiomics} \\
MedSeg Liver Segments \cite{medseg_database}  & 50 & 9 & CT & Liver Segments & \url{https://www.medseg.ai/database/liver-segments-50-cases} \\
MedSeg Vasculature Brain \cite{medseg_database} & 1 & 73 & MRI & Brain Vessels & \url{https://www.medseg.ai/database/brain-vasculature} \\
MedSeg Vasculature Abdomen \cite{medseg_database} & 1 & 43 & CT & Abdominal Vessels & \url{https://www.medseg.ai/database/vasculature-of-the-abdomen} \\
MedSeg Vasculature Neck \cite{medseg_database} & 1 & 19 & CT & Abdominal Vasculature & \url{https://www.medseg.ai/database/vasculature-of-the-neck} \\
MedSeg Vasculature Pelvis \cite{medseg_database} & 1 & 74 & CT & Pelvic Vasculature & \url{https://www.medseg.ai/database/vasculature-of-the-pelvis} \\
MedSeg Musulature Pelvis \cite{medseg_database} & 1 & 67 & CT & Pelvic Musculature & \url{https://www.medseg.ai/database/musculature-of-the-pelvis} \\
MedSeg Brain Ventricle \cite{medseg_database} & 10 & 1 & MRI & Brain Ventricles & \url{https://www.medseg.ai/database/lateral-ventricles-50-mri-cases} \\
\bottomrule
\end{tabular}
}
\caption{\textbf{Overview of training datasets part 2.} Overview of the 158 datasets used for model training (Part 2/2), covering names, image counts, number of categories, modalities, targets, and access links.}
\vspace*{14em}
\label{tab:appendix_train_datasets_2}
\end{table*}

\subsection{Large-Scale Multi-Modality Dataset}
To train VoxTell, we assembled a comprehensive multi-modality 3D medical imaging corpus by aggregating 158 publicly available datasets, encompassing over 62,000 volumetric scans ($\approx$4~TB) and 1,087 distinct anatomical and pathological concepts. The collection spans Computed Tomography (CT), multi-sequence Magnetic Resonance Imaging (MRI), and Positron Emission Tomography (PET), covering both healthy and pathological anatomies across the brain, thorax, abdomen, pelvis, musculoskeletal system, and vasculature.

\noindent Dataset targets range from large organs (\eg liver, lungs, heart, brain) to fine-grained substructures (\eg hippocampal subfields, aortic segments, vertebral bodies) and pathological lesions (\eg tumors, vascular anomalies, white matter hyperintensities). This breadth yields rich semantic diversity suitable for learning language-conditioned representations.\\

\noindent To give a sense of anatomical coverage, the dataset includes comprehensive labeling of brain structures (\eg cortex, white matter, basal ganglia, ventricles), cardiac structures (atria, ventricles, valves), thoracic, abdominal and pelvic organs (lung, liver, pancreas, kidneys, bladder), musculoskeletal structures (muscles, ligaments, bones), and vascular networks (aorta, venous system, pulmonary vessels), along with diverse pathologies including tumors, lesions, and thrombi. An overview of the structures from coarse to fine is shown in Fig.~\ref{fig:dataset_circle}, whereas all full list of all concepts is given in Tab.~\ref{tab:structures}).\\

% \noindent Full details of all utilized datasets, including the number of images, categories, modalities, and access links, are provided in  Tables~\ref{tab:appendix_train_datasets_1}--\ref{tab:appendix_train_datasets_2}. For model development and ablation studies, we reserve 5\% of each dataset for validation to ensure a consistent evaluation split, with all ablations conducted on this held-out portion to maintain test integrity and benchmark against the current SOTA model SAT in a fair setting (we even give SAT the text encoder we used since its better then theirs (Qwen3-Embedding-4B)). \\

\noindent While we cannot publicly share the full dataset due to licensing constraints, comprehensive details for all 158 datasets, including the number of images, categories, imaging modalities, target structures, and access links, are provided in Appendix Tables~\ref{tab:appendix_train_datasets_1}--\ref{tab:appendix_train_datasets_2}. For model development and ablation studies, we reserve 5\% of each dataset as a held-out validation split to ensure consistent and unbiased evaluation. All ablation experiments are performed exclusively on this subset, preserving test integrity. To enable a fair comparison with the current state-of-the-art model SAT, we also benchmark a variant of SAT on this same training and validation split, providing it with the same text encoder we use (Qwen3-Embedding-4B), which outperforms SAT’s original encoder, shown in Tab. \ref{tab:ablation}. This ensures that improvements are attributable to the model rather than differences in text embeddings or training dataset.

\subsection{In-House Radiotherapy Dataset}
\label{sec:in_house_data}

To assess the models capability of interpreting real-world clinical language, we curated an in-house cohort of 203 patients who underwent stereotactic body radiotherapy (SBRT) for either primary or secondary lung tumors at Heidelberg University Hospital. For each patient, planning CT scans were available, acquired either with or without contrast agent. All scans contained clinically approved, expert-annotated gross tumor volume (GTV) contours, which served as the reference segmentations.\\

\noindent In addition to the imaging data, textual descriptions of the target structures were extracted from the corresponding radiology reports, ensuring a one-to-one correspondence between each annotated lesion and its clinical description. These paired image–text samples constitute a multimodal dataset representative of real-world radiotherapy planning workflows. Example CT image slices and corresponding textual descriptions are shown in Fig.~\ref{fig:teaser}d. The textual findings were translated from German into English using a large language model (gpt-oss-120b), ensuring the translations preserve the semantic precision of the original reports. Representative examples include:

\begin{itemize}
    \item “Suspicious for cavitary bronchogenic carcinoma in the right apical segment of the upper lobe.”
    \item “Histologically confirmed adenocarcinoma NSCLC in the right upper lobe with broad-based pleural contact.”
    \item “Peripheral bronchogenic carcinoma in the right lower lobe with pleural contact.”
    \item “Bronchogenic carcinoma in the left upper lobe with fibrotic streaky consolidations.”
    \item “Suspicious for bronchogenic carcinoma in the left apical upper lobe.”
    \item “Round pulmonary nodule in the left upper lobe associated with known squamous cell carcinoma.”
\end{itemize}

\noindent This dataset was held out entirely during training and used exclusively for independent evaluation.

\subsection{Instance-Focused Findings Dataset}
\label{sec:appendix_instance_dataset_creation}

\begin{table*}[t]
\centering
\setlength{\tabcolsep}{4pt}
\renewcommand{\arraystretch}{1.05}
\begin{adjustbox}{max width=\linewidth}
\begin{tabular}{lcccc}
\toprule
\textbf{Dataset} & \textbf{Images} & \textbf{Modality} & \textbf{Target} & \textbf{Link} \\
\midrule
Decathlon Task 3 \cite{antonelli2021medical,simpson2019large}& 131 & CT & Liver Tumor & \url{http://medicaldecathlon.com} \\
Decathlon Task 6 \cite{antonelli2021medical,simpson2019large}& 63 & CT & Lung Lesion & \url{http://medicaldecathlon.com} \\
Decathlon Task 8 \cite{antonelli2021medical,simpson2019large}& 303 & CT & Hepatic Tumor & \url{http://medicaldecathlon.com} \\
LIDC \cite{lidc}& 1010 & CT & Lung Lesion & \url{https://www.cancerimagingarchive.net/collection/lidc-idri} \\
StructSeg Task4~\cite{structseg} & 50 & CT & Lung Cancer & \url{https://structseg2019.grand-challenge.org} \\
COVID-19-20~\cite{roth2022rapid} & 199 & CT & COVID-19 & \url{https://covid-segmentation.grand-challenge.org/COVID-19-20} \\
Atlas Bourgogne~\cite{quinton2023tumour} & 60 & MRI & Liver Tumor & \url{https://atlas-challenge.u-bourgogne.fr } \\
TriALS~\cite{trials_challenge} & 240 & CT & Liver Lesion & \url{https://www.synapse.org/Synapse:syn53285416/wiki/625814} \\
HCC Tace~\cite{hcctace} & 65 & CT & Hepatocellular Carcinoma & \url{https://www.cancerimagingarchive.net/collection/hcc-tace-seg/} \\
RiderLung~\cite{rider_lung} & 58 & CT & Non-Small Cell Lung Carcinoma & \url{https://www.cancerimagingarchive.net/collection/rider-lung-ct/}\\
Colorectal Liver Mets~\cite{colorectal-liver-metastases} & 171 & CT & Colorectal Liver Metastases & \url{https://www.cancerimagingarchive.net/collection/colorectal-liver-metastases} \\
HCC Tace MRI~\cite{hccMRI} & 34 & MRI & Hepatocellular Carcinoma & \url{https://zenodo.org/records/8179129} \\
BrainGammaKnife~\cite{brain-tr-gammaknife} & 76 & MRI & Brain Tumor & \url{https://www.cancerimagingarchive.net/collection/brain-tr-gammaknife} \\
RexGroundingCT Train~\cite{rexgroundingct} & 2,992 & CT & Chest Findings & \url{https://huggingface.co/datasets/rajpurkarlab/ReXGroundingCT}\\
RADCURE-Tumor~\cite{radcure} & 3,199 & CT & Head Neck Tumor & \url{https://www.cancerimagingarchive.net/collection/radcure} \\
MSWAL~\cite{wu2025mswal} & 484 & CT & Abdominal Lesion & \url{https://huggingface.co/datasets/zhaodongwu/MSWAL/tree/main} \\
NSCLC Pleural Effusion~\cite{kiser2020plethora} & 78 & CT & Pleural Effusion & \url{https://www.cancerimagingarchive.net/analysis-result/plethora} \\ 
NSCLC Radiomics~\cite{NSCLC-Radiomics} & 503 & CT & Lung Lesions & \url{https://www.cancerimagingarchive.net/collection/nsclc-radiomics} \\
\midrule
RexGroundingCT Test~\cite{rexgroundingct} & 50 & CT & Chest Findings & \url{https://huggingface.co/datasets/rajpurkarlab/ReXGroundingCT}\\
\bottomrule
\end{tabular}
\end{adjustbox}
\caption{\textbf{Datasets used for instance-specific training and benchmarking on ReXGroundingCT.} The collection spans multiple organs and modalities (CT, MRI), integrating semantic lesion datasets reformulated into instance-level form, location-rich public datasets with spatial metadata (\eg RADCURE, BrainGammaKnife), and the manually annotated ReXGroundingCT benchmark. Together, these sources enable fine-grained, text-conditioned localization of clinically described findings within complex anatomical contexts.}
\label{tab:instance_datasets}
\end{table*}

While the large-scale semantic dataset enables comprehensive anatomical understanding, current models often fail on \emph{localized, instance-specific} or \emph{context-dependent} queries, an issue also noted in prior work, benchmarking models on the ReXGroundingCT dataset~\cite{baharoon2025state}. They demonstrated that no current model can reliably handle truly instance-specific prompts, and even after fine-tuning on the ReXGroundingCT training split, performance remained suboptimal. To address this gap, we curate an \emph{instance-focused dataset} specifically designed to support reasoning over fine-grained, spatially grounded prompts such as
"spiculated tumor in the left lower lobe"
or
"cluster of HCC lesions in Couinaud segment 5".
For fair comparison, we fine-tune both the current state-of-the-art ReXGroundingCT baseline, SAT~\cite{SAT}, and our proposed model on this extended dataset.

\noindent Our instance dataset is constructed through three complementary pathways:\\

1. \textbf{Conversion of semantic lesion datasets to instance-level form:}  
   We reformulate existing semantic lesion segmentation datasets by converting them into \emph{instance annotations} conditioned on anatomical location. Specifically, we employ TotalSegmentator~\cite{wasserthal2023totalsegmentator} to extract lung lobes and liver sub-segments (Couinaud segments), as well as left and right kidney masks, which then serve as contextual anchors for generating localized textual prompts for lung, liver and kidney lesions.\\

2. \textbf{Integration of location-rich public datasets:}  
   We incorporate publicly available TCIA\footnote{\url{https://www.cancerimagingarchive.net/}} datasets across brain~\cite{brain-tr-gammaknife} and head–neck~\cite{radcure} domains that include original DICOM metadata. These metadata elements (\eg study description, series body part, and slice positioning) are reformulated into \emph{structured location prompts} per segmented object.\\

3. \textbf{Linking free-text findings to spatial annotations:}  
   Complementing the above, we leverage the ReXGroundingCT~\cite{rexgroundingct} dataset, a large-scale, manually annotated benchmark providing \emph{pixel-level 3D segmentations} aligned with corresponding \emph{radiology report findings} from the CT-RATE~\cite{ct-rate} corpus. Unlike the above (semi-)synthetic datasets, this dataset enables explicit grounding of free-text clinical descriptions to precise 3D regions. We use the official training and validation splits respectively.\\

\noindent The full list of datasets used is given in Tab. \ref{tab:instance_datasets}. In contrast to standard semantic segmentation datasets,  which assign voxels to pre-defined classes, our instance-focused dataset is designed to facilitate query-driven localization allowing models to identify specific, user-defined findings within complex anatomical contexts. Within the \textit{VoxTell} architecture, this capability is reinforced through \emph{iterative image--text fusion} during training, promoting the emergence of spatially grounded, instance-aware representations. By moving beyond class-level semantics, this complementary dataset supports clinically meaningful spatial reasoning, directly linking descriptive radiology language to voxel-level anatomical understanding.

\begin{table*}[t]
\centering
\setlength{\aboverulesep}{0.8pt} % default is 4pt
\setlength{\belowrulesep}{0.8pt} % default is 4pt
\setlength{\tabcolsep}{5pt} % default is ~6pt
\renewcommand{\arraystretch}{1.05} % smaller spacing
\begin{adjustbox}{max width=\linewidth}
\begin{tabular}{l|ccc|cccccccc}
\toprule
\textbf{Complexity} & \textbf{Dataset} & \textbf{Images} & \textbf{Modality} & \textbf{Prompt} & \textbf{TotalSegmentator} & \textbf{BiomeParse}  & \textbf{Text3DSam} & \textbf{SegVol} & \textbf{BiomedParseV2} & \textbf{SAT} & \textbf{VoxTell}                   \\
\midrule
\multirow{60}{*}{\shortstack{Healthy\\(Known Concept)}}      & \multirow{23}{*}{Pediatric-CT-SEG~\cite{pediatric-ct-seg}} & \multirow{23}{*}{21} & \multirow{23}{*}{CT} & gallbladder                 & 78.91            & 8.74                      & 7.25                      & 71.51                    & 84.63                    & 80.25                     & 85.28                     \\
                                              &                                    &                               &                                & duodenum                    & 53.16            & 2.68                      & 6.47                      & 51.56                    & 59.05                    & 62.17                     & 66.97                     \\
                                              &                                    &                               &                                & right adrenal gland         & 41.27            & 0.38                      & 0.01                      & 37.99                    & 40.52                    & 46.84                     & 48.52                     \\
                                              &                                    &                               &                                & left adrenal gland          & 42.53 & 1.17	& 2.01	& 45.27	& 44.32	& 52.71	 & 52.57                       \\
\bottomrule
\end{tabular}
\end{adjustbox}
\caption{\textbf{Overview of held-out test datasets.} This table presents the full set of evaluation results across all datasets, including healthy and pathological known concepts, as well as generalization to unseen modalities and classes. Dataset metadata is also included. For partially annotated datasets such as Pediatric-CT-SEG and RADCURE-Structures, only the subset containing all annotated classes were used.}
\label{tab:full_results_appendix}
\end{table*}

\section{Vocabulary Construction Details}
\label{sec:appendix_agentic_system}

\begin{figure*}[t]
  \centering
  \includegraphics[width=\textwidth]{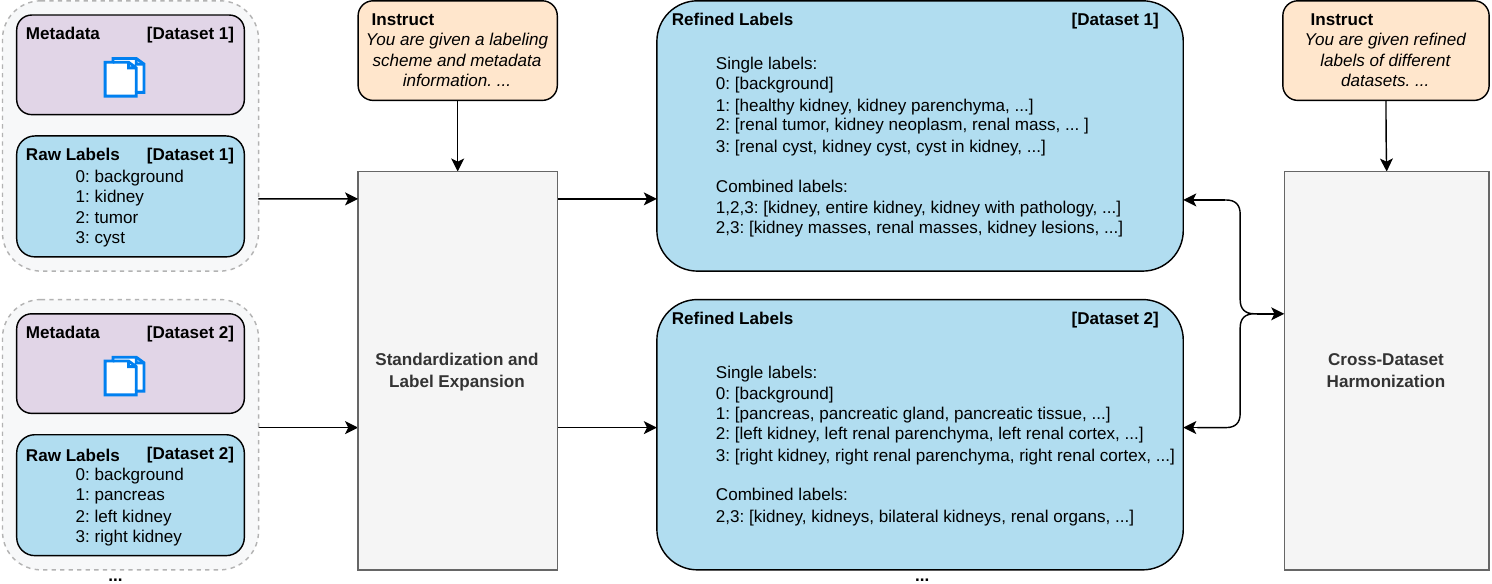}
 \caption{\textbf{Iterative pipeline for large-scale label standardization.}
Each dataset provides raw labels and metadata, which are processed by a \textit{Semantic Label Expansion} module to generate structured and semantically consistent label variants.
A \textit{Cross-Dataset Harmonization} module analyzes these expanded label sets to identify inconsistent or overlapping label definitions and proposes standardized mappings across datasets.
Experts manually review proposed mappings, refine the expanded labels when necessary, and ensure that final label definitions remain faithful to the original dataset semantics.}
  \label{fig:vocab_construction}
\end{figure*}

We construct a unified vocabulary across 158 heterogeneous 3D medical segmentation datasets using an iterative pipeline comprising three components: (1) semantic label expansion, (2) cross-dataset harmonization, and (3) human expert validation. The final vocabulary contains 1{,}087 unified concepts and 9{,}682 rewritten labels. Figure~\ref{fig:vocab_construction} illustrates the pipeline. For the expansion and harmonization stages, we use Anthropic Claude Sonnet 4~\cite{claude_sonnet_4} with extended thinking enabled. \\

1. \textbf{Semantic Label Expansion:}\\
For each dataset, we first consolidate instance-level annotations into semantic base labels (\eg merging individual tumor instances into a single tumor concept). These base labels are then processed together with publicly available metadata, including published dataset papers of the dataset, challenge documentation, dataset websites, and SNOMED CT ontology codes when available. The instruction set in Fig.~\ref{fig:label_generation_prompt} generates:
\begin{itemize}[noitemsep,nolistsep]
    \item single-label alternatives for each semantic base label,
    \item combined-label alternatives for clinically meaningful groupings (\eg combining individual rib labels into "rib cage"),
    \item a minimum of five validated variants per concept.
\end{itemize}

For example, a base label "kidney" is expanded to include variants such as "renal parenchyma", "renal organ", and "kidney tissue". Positional terms (\eg "left", "anterior") must always include the anatomical structure name to ensure self-contained descriptions. Combined labels are generated only for established clinical groupings, such as merging left and right kidney labels into "kidneys" or "bilateral kidneys". Organ labels by default refer to the complete structure including any pathologies unless explicitly specified otherwise (\eg "liver parenchyma only").\\

2. \textbf{Cross-Dataset Harmonization:}\\
Identical label strings may denote different anatomical targets or coverage definitions across datasets. To resolve such inconsistencies, an LLM-based harmonization stage processes expanded label sets together with metadata and ontology-derived descriptors. The conflict-detection prompt in Fig.~\ref{fig:label_harmonization_prompt} outputs structured JSON specifying:
\begin{itemize}[noitemsep,nolistsep]
    \item conflict presence and severity (none / minor / major),
    \item anatomical discrepancy description,
    \item recommended harmonization action,
    \item affected datasets.
\end{itemize}

Major conflicts include differences in organ identity (\eg "left ventricle" referring to myocardial wall versus blood-filled chamber), inclusion of pathologies (\eg "liver" with or without tumors), or whole-organ versus subregion definitions. Minor conflicts reflect boundary variations or substructure inclusion differences. For instance, if "kidney" refers to healthy parenchyma in one dataset but includes tumors and cysts in another, these are maintained as distinct concepts with explicit descriptions. Ontology terms support this process by providing standardized anatomical references.\\

3. \textbf{Human Expert Review:}\\
All major-conflict cases and all final label variants are reviewed by a human expert, which verifies anatomical correctness, alignment with dataset semantics as described in published papers, and consistency within hierarchical groupings (\eg ensuring "thoracic spine" variants align with individual vertebra labels). The reviewer can accept, modify, or reject any LLM-generated suggestion. All modifications are documented to ensure consistency across the final vocabulary. This validation step ensures that the vocabulary remains faithful to established medical terminology and prevents the introduction of spurious or anatomically inaccurate terms.\\

\lstdefinestyle{promptstyle}{
    backgroundcolor=\color{gray!10},
    basicstyle=\fontsize{7}{7}\selectfont\fontfamily{pcr}\selectfont\color{black},
    frame=single,
    rulecolor=\color{gray!50},
    frameround=tttt,
    breaklines=true,
    breakatwhitespace=false,
    columns=flexible,
    keepspaces=true,
    xleftmargin=2mm,
    xrightmargin=2mm,
    framexleftmargin=2mm,
    framexrightmargin=2mm,
    upquote=true,  % Use straight quotes (cleaner for code)
    literate={"}{\textquotedbl}1,  % Or use typographic quotes
}
\begin{figure*}[p]
\centering
\begin{lstlisting}[style=promptstyle]
Analyze if the label '{label_name}' has conflicting definitions across datasets - meaning it refers to different anatomical structures or body parts.
{dataset_definitions}
Focus: Does this label text refer to different parts of the body or include/exclude different anatomical components?
Conflicts to detect:
1. Different anatomical structures entirely (e.g. 'mass' = lung mass in one dataset, liver mass in another)
2. Different inclusion of sub-structures (e.g. 'brain' with CSF vs without CSF)
3. Whole organ vs part of organ (e.g. 'liver' = whole organ vs 'liver' = 'parenchyma only')
4. Different pathology inclusions (e.g. 'kidney' with tumors vs without)
Not conflicts:
- Terminology variations for same structure ('hepatic' vs 'liver')
- Different alternative phrasings in the alternatives list
- Minor wording differences
Output (Json only):
{{
    'has_conflict': true/false,
    'conflict_severity': 'none'/'minor'/'major',
    'conflict_description': 'What anatomical difference exists, or 'No conflicts'',
    'recommendations': ['action 1', 'action 2'],
    'affected_datasets': ['Dataset001', ...]
}}
Severity:
- 'none': Same anatomical structure across all datasets
- 'minor': Small inclusion differences (e.g. organ edge definitions)
- 'major': Different anatomical structures or major component differences
JSON only, no other text.
\end{lstlisting}
\caption{\textbf{Instructions for cross-dataset conflict detection.}
Label definitions are analyzed using these instructions to identify cases where identical text refers to different anatomical structures across datasets.
The instructions distinguish genuine semantic conflicts (\eg different organs, inclusion/exclusion of pathologies) from benign terminology variations.
The structured output provides conflict severity assessment and resolution recommendations for systematic harmonization before multi-dataset training.}
\label{fig:label_harmonization_prompt}
\end{figure*}

\lstdefinestyle{promptstyle}{
    backgroundcolor=\color{gray!10},
    basicstyle=\fontsize{7}{7}\selectfont\fontfamily{pcr}\selectfont\color{black},
    frame=single,
    rulecolor=\color{gray!50},
    frameround=tttt,
    breaklines=true,
    breakatwhitespace=false,
    columns=flexible,
    keepspaces=true,
    xleftmargin=2mm,
    xrightmargin=2mm,
    framexleftmargin=2mm,
    framexrightmargin=2mm,
    upquote=true,  % Use straight quotes (cleaner for code)
    literate={"}{\textquotedbl}1,  % Or use typographic quotes
}
\begin{figure*}[p]
\centering
\begin{lstlisting}[style=promptstyle]
You are given a labeling scheme and metadata information.
The labeling has a format like the following example:

'labels': {
        'background': 0,
        'liver tumor': 1,
        'liver': [1, 2],
        ...
    },
...

where a text label is mapped to one or to a list of label ids.
Given the segmentation dataset metadata and label definitions, generate comprehensive and diverse alternative labels for every segmentation label in the dataset.json in a SINGLE, complete JSON response.
IMPORTANT: The output JSON must use numerical label IDs as keys, not the original label names. Map each ID to a list of alternative text labels.
The output format must be a single, complete JSON object:
{'single_labels': {
    '0': ['background'],
    '1': ['alternative_name_1', 'alternative_name_2', ...],
    '2': ['alternative_name_1', 'alternative_name_2', ...],
    ...
},
'combined_labels': {
    '1,2': ['combined_name_1', 'combined_name_2', ...],
    // Include meaningful combinations using numeric IDs
}}

Use the given information in the metadata to diversity the labels in a useful manner.
Critical for multi-dataset training:
1. Every alternative label must be fully self-contained and unambiguous across datasets
2. Positional terms (e.g. 'anterior', 'left') must never be used in isolation-always pair them with the anatomical structure (e.g. 'anterior hippocampus')
3. Always include the anatomical/structural name in every alternative (e.g. 'anterior hippocampus' not just 'anterior')
4. For subregions or subdivisions, always maintain the parent structure in the label (e.g. 'hippocampal head' NOT just 'head')
5. Any abbreviation must be domain-specific and widely recognized (e.g. 'L. hippocampus' for left hippocampus is acceptable)
6. Labels must be uniquely identifiable even when used across multiple different dataset contexts
7. Don't assign duplicate labels to different structures or combinations (e.g. assigning 'liver' to both '2' and '1,2' is not allowed)
8. Organ labels always mean the whole organ including its components (e.g. lesions, tumors, vessels)
9. Critical: For grouped instances (e.g. 'mediastinal lymph nodes': [1, 2, 3, ...]):
   - Do not create individual entries for these IDs in 'single_labels'
   - Instead, add them directly to 'combined_labels' with the comma-separated IDs as the key
   - Example: For {'mediastinal lymph nodes': [1, 2, 3, ..., 56]}, add '1,2,3,...,56': ['mediastinal lymph nodes', ...] to 'combined_labels'
   - These IDs should not appear individually in 'single_labels'
Diversity requirements:
1. For each label, provide at least 5 diverse yet clinically accurate alternatives
2. Include variations in terminology across different medical contexts:
   - Formal anatomical terminology (e.g. 'hepatic tissue')
   - Clinical shorthand used in practice (e.g. 'HCC' for 'hepatocellular carcinoma')
   - Descriptive terms used in radiology reports
   - Common terms used when discussing with patients
3. Consider different ways to describe the same structure:
   - Positional descriptions with structure names (e.g. 'superior cerebellar peduncle', not just 'superior')
   - Functional descriptions where relevant, always with the structure name
   - Size/shape-based descriptions where appropriate, always with the structure name
Quality control:
1. Every alternative must be clinically accurate and actually used in medical contexts
2. Avoid artificially creating diversity by using uncommon or imprecise terminology
3. Only include terms that medical professionals would recognize and use
4. Maintain anatomical precision - alternatives must refer to the exact same structure
5. For combined labels, only create meaningful clinical groupings that would be referred to together. Do not combine labels with the background label.
6. Do not create alternatives for the background label.
Consistency guidelines:
1. Maintain parallel structure across related anatomical terms
2. Use consistent terminology for laterality (left/right) and positioning
3. When creating alternatives for a series of related structures, ensure naming patterns are consistent
4. Unqualified organ names (e.g. 'liver') naturally refer to the complete organ including any pathologies, unless explicitly specified otherwise

Verification: 
Before returning your response:
- Ensure every label ID from dataset.json is represented in single_labels
- Verify that combined_labels ONLY reference label IDs that actually exist in single_labels
- Check that no non-existent label IDs are included in any combined label keys
- Confirm that each label has at least 5 diverse alternatives
- Verify that all alternatives accurately represent the original label
Only return the json, nothing else.
\end{lstlisting}
\caption{\textbf{Instructions for label standardization and expansion.}
Each dataset's raw label definitions and metadata are processed using these instructions to generate a standardized JSON output containing: (1) diverse alternative labels for each label ID, ensuring anatomical precision and self-contained descriptions (\eg "anterior hippocampus" not "anterior"), and (2) meaningful combined labels representing valid anatomical groupings.
The instructions enforce at least 5 clinically accurate alternatives per label and prevent cross-dataset ambiguity to support robust text-conditioned segmentation.}
\label{fig:label_generation_prompt}
\end{figure*}

\section{Text Embedding Model Selection}
\label{sec:appendix_text_encoder_selection}

\begin{table}[t]
\centering
\setlength{\tabcolsep}{4pt}
\renewcommand{\arraystretch}{1.05}
\begin{adjustbox}{max width=\columnwidth}
\begin{tabular}{lcc}
\toprule
\textbf{Text Embedding Model} & \textbf{Model Size} & \textbf{Dice} \\
\midrule
EmbeddingGemma \cite{embedding_gemma_2025} & 300M & 56.77 \\
\textbf{SAT Text Encoder} \cite{SAT} & 450M & 60.43 \\
\rowcolor{gray!10}
Qwen3-Embedding-0.6B \cite{qwen3embedding} & 600M & \textbf{62.24} \\
\midrule
\rowcolor{gray!10}
Qwen3-Embedding-4B \cite{qwen3embedding} & 4B & \textbf{62.55} \\
\midrule
Jasper En Vision-Language v1 \cite{zhang2025jasperstelladistillationsota} & 7B & 58.42 \\
SFR-Embedding-2 \cite{SFR-embedding-2} & 7B & 62.08 \\
GTE-Qwen2-7B-instruct \cite{gte-Qwen2-7B-instructli2023towards} & 7B & 62.44 \\
Linq-Embed-Mistral \cite{LinqAIResearch2024} & 7B & 62.78 \\
E5-Mistral-7b-instruct \cite{e5-mistral-7b-instructwang2022text, e5-mistral-7b-instructwang2023improving} & 7B & \textbf{62.88} \\
\rowcolor{gray!10}
Qwen3-Embedding-8B \cite{qwen3embedding} & 8B & 62.45 \\
\bottomrule
\end{tabular}
\end{adjustbox}
\caption{\textbf{Comparison of text embedding models.}
We evaluate the top open-source text encoders from the MTEB benchmark~\cite{muennighoff2022mteb} alongside the \textbf{SAT Text Encoder}~\cite{SAT}, using identical training and validation data. Models are grouped by size (small, medium, large). Across all groups, \fcolorbox{gray!10}{gray!10}{\textbf{Qwen3-Embedding}} shows consistently strong performance, achieving the highest overall Dice. Performance saturates below 63\%, so we adopt the medium-scale variant to balance accuracy and computational efficiency.}
\label{tab:text_encoder_ablation}
\end{table}

To identify an effective text encoder for our framework, we evaluate several state-of-the-art text embedding models from the MTEB\footnote{\url{http://mteb-leaderboard.hf.space/?benchmark_name=MTEB\%28Medical\%2C+v1\%29}} benchmark under consistent training and validation conditions. Given the large number of parameters in most text encoders and to preserve their pretrained knowledge, we freeze all encoder weights during training to efficiently apply it in downstream tasks. Candidate models are grouped by scale: small (300M–600M), medium (4B), and large (7B–8B). All encoders remain frozen during training, and instruction-tuned variants follow a uniform prompt:
\textit{``Instruct: Given an anatomical term query, retrieve the precise anatomical entity and location it represents. Query: [anatomical term].''}\\

\noindent As shown in Table~\ref{tab:text_encoder_ablation}, segmentation performance generally improves with model capacity. Small-scale models such as EmbeddingGemma and SAT Text Encoder exhibit noticeably lower Dice scores, while large-scale encoders (7B–8B) yield strong results but demand more than 24 GB of GPU memory, often far exceeding the capabilities of standard hospital hardware. Among medium-scale candidates, \textbf{Qwen3-Embedding-4B} attains performance comparable to the best large models, offering an optimal balance between accuracy and computational feasibility. Consequently, we adopt Qwen3-Embedding-4B as the default text encoder in VoxTell.

\section{Extended Results}
\label{sec:appendix_limitations}

\begin{table*}[t]
\setlength{\tabcolsep}{4pt}
\renewcommand{\arraystretch}{1.05}
\begin{adjustbox}{max width=\textwidth}
\begin{tabular}{l|ccccccc}
\toprule
Structure     & \textbf{Patellar Cartilage} & \textbf{Femoral Cartilage} & \textbf{Lateral Tibial Cartilage} & \textbf{Medial Meniscus} & \textbf{Lateral Meniscus} & \textbf{Medial Tibial Cartilage} & \textbf{Mean Dice} \\
\midrule
BioMedParse   & 0.44                                   & 0.54                                  & 0.22                                         & 0.47                                & 0.72                                 & 0.29                                        & 0.45                     \\
Text3DSam     & 0.0                                    & 0.08                                  & 0.09                                         & 0.97                                & 0.1                                  & 0.31                                        & 0.26                     \\
SegVol        & 0.0                                    & 0.0                                   & 0.0                                          & 0.0                                 & 0.0                                  & 0.0                                         & 0.0                      \\
BioMedParseV2 & 0.0                                    & 0.15                                  & 0.0                                          & 0.0                                 & 0.0                                  & 0.0                                         & 0.02                     \\
SAT           & 0.01                                   & 1.34                                  & 0.02                                         & 0.0                                 & 0.0                                  & 0.0                                         & 0.23                     \\
VoxTell       & 0.07                                   & 46.45                                 & 2.97                                         & 0.04                                & 0.0                                  & 14.78                                       & 10.72 \\
\bottomrule
\end{tabular}
\end{adjustbox}
\caption{\textbf{Limitations of zero-shot segmentation on rare body regions.} This evaluation highlights VoxTell and prior text-promptable models on knee MRI (SKM-TEA dataset~\cite{stanford_knee}), a modality and body region very rarely seen during training. Most prompts correspond to fully out-of-distribution (OOD) anatomical structures, making it extremely challenging for the model to interpolate. As shown, VoxTell achieves partial recognition of some structures but, like other methods, struggles to generalize to entirely novel spatial and visual patterns, illustrating the limitations discussed in the main paper.}
\label{tab:limitations}
\end{table*}

\paragraph{Results per prompt}
Table~\ref{tab:full_results_appendix} presents a comprehensive overview of segmentation performance across all held-out test datasets, extending the results reported in the main paper (Tab.~\ref{tab:model_comparison}). For healthy anatomical structures with known concepts, such as those in Pediatric-CT-SEG, AeroPath, VEELA2025, RADCURE, and HVSMR-2.0, VoxTell consistently achieves the highest Dice scores across the majority of organs and structures, often substantially outperforming prior methods such as TotalSegmentator, BiomeParse, Text3DSam, SegVol, BiomedParseV2, and SAT. Notably, VoxTell maintains strong performance on challenging, small, or laterality-sensitive structures such as adrenal glands, breast tissue, and the spinal canal, where competing methods frequently fail or produce near-zero predictions.\\

\noindent On pathological datasets with known concepts, including ARCIN NSCLC, ISBI MS, Adrenal-ACC-Ki67, HCC-TACE-SEG, Pengwin, and BrainMetShare, VoxTell demonstrates robust segmentation of tumors and lesions across CT, PET and MR modalities, achieving substantial improvements over previous approaches that often struggle with heterogeneous appearances or low contrast. In cross-modality scenarios, such as QIN Breast, PANTHER Task 1, and Soft-Tissue-Sarcoma, VoxTell exhibits superior generalization, successfully segmenting structures unseen during training despite modality shifts. Finally, for unknown concepts, including FedBCa and MedSeg Esophageal, VoxTell significantly outperforms all baseline methods, demonstrating its capacity for zero-shot anatomical understanding and reliable localization of rare or unseen structures. Overall, these extended results confirm that VoxTell achieves state-of-the-art performance across a diverse range of anatomical, pathological, and cross-modal segmentation tasks, highlighting its robustness, generalization, and practical applicability in clinical imaging.\\

\paragraph{Zero-shot segmentation on rare body regions}
To probe the limits of open-vocabulary generalization, we evaluated VoxTell and prior text-promptable segmentation models on the Stanford Knee MRI dataset~\cite{stanford_knee}. The knee region is sparsely represented in the training set and includes cartilage and meniscus structures absent from training data. As Table~\ref{tab:limitations} shows, all models struggle on these highly out-of-distribution (OOD) structures, which differ both spatially and visually from familiar anatomy, though VoxTell is able to segment a few structures with lower performance.\\

\noindent Despite low absolute Dice scores, VoxTell partially recognizes key anatomical components such as femoral and tibial cartilage. This demonstrates the frontier of zero-shot medical segmentation: while VoxTell generalizes across modalities and related anatomical concepts (Tab.~\ref{tab:ood}), completely unseen regions remain challenging. These findings suggest a promising path for combining vision–language pretraining with lightweight few-shot or report-level fine-tuning to close remaining gaps.

\clearpage

\onecolumn
\begin{longtable}{p{2cm}|p{3.5cm}|p{11.35cm}}
\toprule
\textbf{Region} & \textbf{Subregion} & \textbf{Structures (Count)} \\
\midrule
\endfirsthead
\multicolumn{3}{c}{\textit{(continued from previous page)}} \\
\toprule
\textbf{Region} & \textbf{Subregion} & \textbf{Structures (Count)} \\
\midrule
\endhead
%\midrule
\multicolumn{3}{r}{\textit{(continued on next page)}} \\
\endfoot
\bottomrule
\caption{\textbf{Summary of anatomical structures.} 
Overview of all 1078 distinct anatomical labels from public training datasets. 
Each entry lists the structure name and its training case frequency (in parentheses), organized by anatomical region and subregion.
}
\label{tab:structures}
\endlastfoot
Brain & Cerebral structures & brainstem (7360), brain (4330), left cerebral white matter (3654), right cerebral white matter (3654), left cerebral cortex (3644), right cerebral cortex (3644), white matter (519), gray matter (489), white matter hyperintensities (60), deep gray matter (40), cortical gray matter (30), white matter lesions (30) \\
\cline{2-3}
 & Ventricles and csf & cerebrospinal fluid (4133), left lateral ventricle (3654), right lateral ventricle (3654), left inferior lateral ventricle (3644), right inferior lateral ventricle (3644), 3rd ventricle (3644), 4th ventricle (3644), left brain lateral ventricle (158), right brain lateral ventricle (158), lateral ventricles (80), third ventricle (10), fourth ventricle (10) \\
\cline{2-3}
 & Basal ganglia & left caudate (3654), right caudate (3654), left putamen (3654), right putamen (3654), left pallidum (3654), right pallidum (3654), left accumbens area (3644), right accumbens area (3644), left caudate nucleus (158), right caudate nucleus (158), left lentiform nucleus (158), right lentiform nucleus (158), basal ganglia (30), left nucleus accumbens (10), right nucleus accumbens (10) \\
\cline{2-3}
 & Thalamus and diencephalon & left thalamus (3654), right thalamus (3654), left ventral dc (3644), right ventral dc (3644), left ventral diencephalon (10), right ventral diencephalon (10) \\
\cline{2-3}
 & Hippocampus & left hippocampus (3932), right hippocampus (3932), anterior hippocampus (260), posterior hippocampus (260), left hippocampus subiculum (20), left hippocampus cornu ammonis 2 (20), left hippocampus cornu ammonis 1 (20), left hippocampus cornu ammonis 4 and gyrus dentatus (20), left hippocampus entorhinal cortex (20), left hippocampus cornu ammonis 3 (20), left hippocampus tail (20), right hippocampus subiculum (20), right hippocampus cornu ammonis 2 (20), right hippocampus cornu ammonis 1 (20), right hippocampus cornu ammonis 4 and gyrus dentatus (20), right hippocampus entorhinal cortex (20), right hippocampus cornu ammonis 3 (20), right hippocampus tail (20) \\
\cline{2-3}
 & Amygdala and limbic & left amygdala (3654), right amygdala (3654), corpus callosum (10) \\
\cline{2-3}
 & Cerebellum & left cerebellum white matter (3644), right cerebellum white matter (3644), left cerebellum cortex (3644), right cerebellum cortex (3644), cerebellum (70), cerebellar vermis superior posterior (10), cerebellar vermis anterior (10), cerebellar vermis inferior posterior (10), left cerebellar hemisphere (10), right cerebellar hemisphere (10) \\
\cline{2-3}
 & Frontal lobe & left caudal anterior cingulate (10), left caudal middle frontal (10), left lateral orbitofrontal (10), left medial orbitofrontal (10), left paracentral lobule (10), left pars opercularis (10), left pars orbitalis (10), left pars triangularis (10), left precentral gyrus (10), left rostral anterior cingulate (10), left rostral middle frontal (10), left superior frontal (10), left frontal pole (10), right caudal anterior cingulate (10), right caudal middle frontal (10), right lateral orbitofrontal (10), right medial orbitofrontal (10), right paracentral lobule (10), right pars opercularis (10), right pars orbitalis (10), right pars triangularis (10), right precentral gyrus (10), right rostral anterior cingulate (10), right rostral middle frontal (10), right superior frontal (10), right frontal pole (10) \\
\cline{2-3}
 & Temporal lobe & left temporal lobe (120), right temporal lobe (120), left temporal lobes (50), right temporal lobes (50), left banks superior temporal sulcus (10), left entorhinal (10), left fusiform gyrus (10), left inferior temporal (10), left middle temporal (10), left parahippocampal gyrus (10), left superior temporal (10), left temporal pole (10), left transverse temporal (10), right banks superior temporal sulcus (10), right entorhinal (10), right fusiform gyrus (10), right inferior temporal (10), right middle temporal (10), right parahippocampal gyrus (10), right superior temporal (10), right temporal pole (10), right transverse temporal (10) \\
\cline{2-3}
 & Parietal lobe & left inferior parietal (10), left postcentral gyrus (10), left posterior cingulate (10), left precuneus (10), left superior parietal (10), left supramarginal gyrus (10), right inferior parietal (10), right postcentral gyrus (10), right posterior cingulate (10), right precuneus (10), right superior parietal (10), right supramarginal gyrus (10) \\
\cline{2-3}
 & Occipital lobe & left cuneus (10), left lateral occipital (10), left lingual gyrus (10), left pericalcarine (10), right cuneus (10), right lateral occipital (10), right lingual gyrus (10), right pericalcarine (10) \\
\cline{2-3}
 & Insula and cingulate & left isthmus cingulate (10), left insula (10), right isthmus cingulate (10), right insula (10) \\
\cline{2-3}
 & Other cortical & extra-axial space (40), left unknown cortex (10), right unknown cortex (10) \\
\cline{2-3}
 & Anterior circulation arteries & right internal carotid artery (833), left internal carotid artery (833), anterior communicating artery (300), right anterior cerebral artery (250), left anterior cerebral artery (250), right middle cerebral artery (250), left middle cerebral artery (250), third segment of anterior cerebral artery (a2) (250), right a1 and a2 of anterior cerebral artery (50), left a1 and a2 of anterior cerebral artery (50), right a3 of anterior cerebral artery (50), left a3 of anterior cerebral artery (50), third a2 of anterior cerebral artery (50), third a3 of anterior cerebral artery (50), right m1 of middle cerebral artery (50), left m1 of middle cerebral artery (50), right m2 of middle cerebral artery (50), right m3 of middle cerebral artery (50), left m2 of middle cerebral artery (50), left m3 of middle cerebral artery (50), paraopthalmic part of right internal carotid artery (1), posterior communicating part of right internal carotid artery (1), anterior choroidal part of the right internal carotid artery (1), right carotid terminus (1), right m1 (1), right a1 (1), anterior communicating artery partially fenestrated (1), right a2 (1), left paraopthalmic part of the internal carotid (1), left posterior communicating part of the internal carotid (1), left anterior choroidal artery part of the internal carotid (1), left carotid terminus (1), left a1 (1), left a2 (1), left m1 (1), insular segment of the right middle cerebral artery (1), insular branches of the left middle cerebral artery (m3/m4) (1), right anterior insular part of middle cerebral artery (m3/m4) (1), right posterior insular part of the middle cerebral artery (1), right superior terminal branch of the middle cerebral artery (1), right parietal cortical branches of the insular part of the middle cerebral artery (1), left m2-branches (1), left m3/m4 (1) \\
\cline{2-3}
 & Posterior circulation arteries & basilar artery (302), right posterior cerebral artery (250), left posterior cerebral artery (250), right superior cerebellar artery (51), left superior cerebellar artery (51), right vertebral artery (50), left vertebral artery (50), right p1 and p2 of posterior cerebral artery (50), left p1 and p2 of posterior cerebral artery (50), right p3 and p4 of posterior cerebral artery (50), left p3 and p4 of posterior cerebral artery (50), right anterior inferior cerebellar artery (50), left anterior inferior cerebellar artery (50), right posterior inferior cerebellar artery (50), left posterior inferior cerebellar artery (50), right p1 (1), right p2 (1), left p1 (1), left p2 (1), left p3 (1), left p4/p5 (1) \\
\cline{2-3}
 & Circle of willis & right posterior communicating artery (301), left posterior communicating artery (301), right anterior choroidal artery (51), left anterior choroidal artery (51) \\
\cline{2-3}
 & Small cerebral arteries & right ophthalmic artery (50), left ophthalmic artery (50), left opthalmic artery (1), right opthalmic artery (1), right lenticolstriate arteries (1), left lenticulostriate arteries (1), right posteromedial central arteries (1), left posteromedial central arteries (1), right paramedian artery branch (1), right medial frontobasal artery (1), right lateral frontobasal artery (1), left medial orbitofrontal artery (1), left polar frontal artery (1), right polar frontal artery (1), left pericallosal artery (1), left callosomarginal artery (a3/4) (1), right callosomarginal artery (a3/4) (1), right anterior internal frontal artery (1), left anterior internal frontal artery (1), right intermediate internal frontal artery (1), left intermediate internal frontal artery (1), right prefrontal artery (1), right artery of precentral gyrus (1), right artery of central sulcus (1), right polar temporal artery (1), right temporal branch (1), right anterior temporal artery branch (1), right middle temporal artery branch (1), right posterior parietal artery (1), right posterior temporal artery branch of the mca (1), right parietal branches of the middle cerebral artery (1), right posterior temporal branches of the pca (1), right medial occipital artery (1), right precuneal branches (1), right calcarine branches (1), right parietooccipital branches (1) \\
\cline{2-3}
 & Cerebral veins and sinuses & superior sagittal sinus (26), vein of galen (25), straight sinus (25), internal cerebral veins (25), right basal vein of rosenthal (25), left basal vein of rosenthal (25) \\
\cline{2-3}
 & Meningeal vessels & right middle meningeal artery (26), left middle meningeal artery (26) \\
\cline{2-3}
 & Brain pathology & enhancing tumor (4681), edema (2840), necrosis/non-enhancing tumor core (2183), non-enhancing tumor core (1943), surrounding non-enhancing hyperintensity (1943), resection cavity (1943), brain lesion (655), gross tumor volume (500), stroke (453), multiple sclerosis lesion (366), necrotic tumor (294), non-enhancing tumor (261), cystic component (261), brain tumor (186), hie lesion (85), cerebral microbleed (72), infarction (30) \\
\midrule
Head and neck & Eyes and orbits & left eye (3505), right eye (3505), left lens (3505), right lens (3505), left eyeball (533), right eyeball (533), eyeball globe (400), orbital fat (400), left anterior eyeball (84), right anterior eyeball (84), left posterior eyeball (84), right posterior eyeball (84) \\
\cline{2-3}
 & Extraocular muscles & eye superior rectus muscle (400), eye lateral rectus muscle (400), eye medial rectus muscle (400), eye inferior rectus muscle (400), eye superior oblique muscle (400) \\
\cline{2-3}
 & Optic pathway & left optic nerve (3636), right optic nerve (3636), chiasma opticum (3636), optic nerve (400) \\
\cline{2-3}
 & Lacrimal system & left lacrimal gland (3419), right lacrimal gland (3419), lacrimal gland (400) \\
\cline{2-3}
 & Facial structures & left cheek (533), right cheek (533) \\
\cline{2-3}
 & Inner ear & left cochlea (3539), right cochlea (3539), cochlea (210), left vestibular semicircular canal (120), right vestibular semicircular canal (120), left inner ear (50), right inner ear (50) \\
\cline{2-3}
 & Middle ear & left middle ear (170), right middle ear (170), left tympanic cavity (120), right tympanic cavity (120), left eustachian tube bone (120), right eustachian tube bone (120) \\
\cline{2-3}
 & Temporal bone & left internal auditory canal (120), right internal auditory canal (120), left mastoid process (120), right mastoid process (120) \\
\cline{2-3}
 & Cranial bones & mandible (3466), skull (3358), lower jawbone (532), upper jawbone (532), left inferior alveolar canal (532), right inferior alveolar canal (532), left mandibular incisive canal (532), right mandibular incisive canal (532), lingual canal (532), left mandible (170), right mandible (170), left temporomandibular joint (170), right temporomandibular joint (170) \\
\cline{2-3}
 & Paranasal sinuses & nasal cavity (533), left maxillary sinus (532), right maxillary sinus (532) \\
\cline{2-3}
 & Upper teeth & upper right central incisor (532), upper right lateral incisor (532), upper right canine (532), upper right first premolar (532), upper right second premolar (532), upper right first molar (532), upper right second molar (532), upper right third molar (wisdom tooth) (532), upper left central incisor (532), upper left lateral incisor (532), upper left canine (532), upper left first premolar (532), upper left second premolar (532), upper left first molar (532), upper left second molar (532), upper left third molar (wisdom tooth) (532) \\
\cline{2-3}
 & Lower teeth & lower left central incisor (532), lower left lateral incisor (532), lower left canine (532), lower left first premolar (532), lower left second premolar (532), lower left first molar (532), lower left second molar (532), lower left third molar (wisdom tooth) (532), lower right central incisor (532), lower right lateral incisor (532), lower right canine (532), lower right first premolar (532), lower right second premolar (532), lower right first molar (532), lower right second molar (532), lower right third molar (wisdom tooth) (532) \\
\cline{2-3}
 & Tooth pulp & upper right central incisor pulp (532), upper right lateral incisor pulp (532), upper right canine pulp (532), upper right first premolar pulp (532), upper right second premolar pulp (532), upper right first molar pulp (532), upper right second molar pulp (532), upper right third molar (wisdom tooth) pulp (532), upper left central incisor pulp (532), upper left lateral incisor pulp (532), upper left canine pulp (532), upper left first premolar pulp (532), upper left second premolar pulp (532), upper left first molar pulp (532), upper left second molar pulp (532), upper left third molar (wisdom tooth) pulp (532), lower left central incisor pulp (532), lower left lateral incisor pulp (532), lower left canine pulp (532), lower left first premolar pulp (532), lower left second premolar pulp (532), lower left first molar pulp (532), lower left second molar pulp (532), lower left third molar (wisdom tooth) pulp (532), lower right central incisor pulp (532), lower right lateral incisor pulp (532), lower right canine pulp (532), lower right first premolar pulp (532), lower right second premolar pulp (532), lower right first molar pulp (532), lower right second molar pulp (532), lower right third molar (wisdom tooth) pulp (532) \\
\cline{2-3}
 & Dental prosthetics & bridge (532), crown (532), implant (532) \\
\cline{2-3}
 & Oral soft tissues & lips (3419), oral cavity (204), buccal mucosa (84) \\
\cline{2-3}
 & Salivary glands & left parotid gland (3636), right parotid gland (3636), left submandibular gland (3586), right submandibular gland (3586) \\
\cline{2-3}
 & Thyroid and pituitary & left thyroid (2226), right thyroid (2226), thyroid gland (1446), pituitary gland (254) \\
\cline{2-3}
 & Pharynx & pharynx (532), pharyngeal constrictor muscle (120), nasopharynx cancer (50) \\
\cline{2-3}
 & Larynx & larynx (3455), larynx glottis (204), larynx supraglottic (204), arytenoid cartilage (84), cricopharyngeal inlet (84) \\
\cline{2-3}
 & Carotid system & left common carotid artery (1931), right common carotid artery (1761), left carotid artery (98), right carotid artery (98), right external carotid artery (25), left external carotid artery (25), right superficial temporal artery (25), left superficial temporal artery (25), right maxillary artery (25), left maxillary artery (25), common carotid artery (1), internal carotid artery (1), external carotid artery (1) \\
\cline{2-3}
 & Jugular venous system & right internal jugular vein (533), left internal jugular vein (533), internal jugular vein (1), external jugular vein (1), posterior auricular vein (1), retromandibular vein (posterior branch) (1), facial vein (1), transverse cervical veins (1), superficial temporal veins (1), middle temporal vein (1), deep superficial veins (1), retromandibular vein (anterior branch) (1) \\
\cline{2-3}
 & Subclavian system & brachiocephalic vein (1), subclavian vein (1) \\
\cline{2-3}
 & Other neck structures & head neck lymph node (245) \\
\cline{2-3}
 & Head neck pathology & head neck primary tumor (674), head neck metastatic lymph nodes (674), head neck lesion (245), vestibular schwannoma (210) \\
\midrule
Chest & Lung & left lung (4404), right lung (4404), lung (3427) \\
\cline{2-3}
 & Lung lobes & left lung lower lobe (3454), right lung lower lobe (3454), right lung middle lobe (3454), left lung upper lobe (3454), right lung upper lobe (3454) \\
\cline{2-3}
 & Airways & trachea (2018), bronchi (583), airways (162) \\
\cline{2-3}
 & Ribs left & 1st left rib (4108), 2nd left rib (4108), 3rd left rib (4108), 4th left rib (4108), 5th left rib (4108), 6th left rib (4108), 7th left rib (4108), 8th left rib (4108), 9th left rib (4108), 10th left rib (4108), 11th left rib (4108), 12th left rib (4108), 13th left rib (1597) \\
\cline{2-3}
 & Ribs right & 1st right rib (4108), 2nd right rib (4108), 3rd right rib (4108), 4th right rib (4108), 5th right rib (4108), 6th right rib (4108), 7th right rib (4108), 8th right rib (4108), 9th right rib (4108), 10th right rib (4108), 11th right rib (4108), 12th right rib (4108), 13th right rib (1597), ribs (654) \\
\cline{2-3}
 & Sternum and clavicles & left clavicula (3974), right clavicula (3974), sternum (3454), rib cartilage (1761), sternum manubrium (533) \\
\cline{2-3}
 & Scapulae & left scapula (3974), right scapula (3974) \\
\cline{2-3}
 & Breast & left breast (1086), right breast (1086) \\
\cline{2-3}
 & Breast pathology & breast lesion (1506) \\
\cline{2-3}
 & Mediastinum & thymus (886), mediastinal tissue (533), mediastinal lymph node (393), mediastinal lymph nodes (192) \\
\cline{2-3}
 & Esophagus & esophagus (7549), cervical esophagus (84) \\
\cline{2-3}
 & Thoracic aorta & left subclavian artery (1931), brachiocephalic artery (1812), right subclavian artery (1761), subclavian artery (1) \\
\cline{2-3}
 & Thoracic veins & left brachiocephalic vein (1761), right brachiocephalic vein (1761), brachiocephalic veins (15), subclavian and carotid arteries (15), azygos (15), azygos vein (2), vertebral artery (1) \\
\cline{2-3}
 & Pleura & pleura lesion (245), pleural effusion (78) \\
\cline{2-3}
 & Lung pathology & lung lesion (1075), lung nodule (481), covid (209), lung tumor (113), non-small cell lung cancer (78), fdg-avid non-small cell lung cancer (69), fdg-avid lung adenocarcinoma (5) \\
\midrule
Heart & Heart & heart (2855), heart tissue (533) \\
\cline{2-3}
 & Atria & left atrium (2516), right atrium (2386), left atrial appendage (1228) \\
\cline{2-3}
 & Ventricles & right ventricle (2800), left ventricular cavity (2503), left ventricle (703), right ventricular cavity (279) \\
\cline{2-3}
 & Atrial walls & left atrium wall (154), left atrium endocardium (154), atrium wall (10) \\
\cline{2-3}
 & Ventricular walls & left ventricle wall (3234) \\
\cline{2-3}
 & Myocardium & left ventricular myocardium (849), myocardium (700), excised myocardium (27) \\
\cline{2-3}
 & Valves & aortic valve (27), mitral valve (27) \\
\cline{2-3}
 & Coronary arteries & coronary artery (1000), left coronary artery (30), right coronary artery (30) \\
\cline{2-3}
 & Pulmonary circulation & pulmonary artery trunk (1713), pulmonary vein (1349), pulmonary artery (593), pulmonary artery tree (316), left atrium and pulmonary veins (137), pulmonary arterial tree (15) \\
\cline{2-3}
 & Systemic cardiac vessels & superior vena cava (1289), aorta with arch branches (110), aorta with branches and coronary arteries (27) \\
\cline{2-3}
 & Cardiac pathology & myocardial infarction (100), no reflow (100), left atrial scars (60), calcification in left anterior descending (32), calcification in left circumflex (32), calcification in right coronary artery (32) \\
\midrule
Abdomen & Liver parenchyma & liver (8669), liver without tumor or vessels (20) \\
\cline{2-3}
 & Liver segments & liver segment 1 (243), liver segment 2 (243), liver segment 3 (243), liver segment 5 (243), liver segment 6 (243), liver segment 7 (243), liver segment 8 (243), liver segment 4 (193), liver segment 4a (50), liver segment 4b (50) \\
\cline{2-3}
 & Liver pathology & liver lesion (4469), hepatic tumor (303), hepatocellular carcinoma (179), colorectal liver metastases (171), liver cancer (131), liver tumor (114), liver cyst (20) \\
\cline{2-3}
 & Pancreas & pancreas (7934), pancreas parenchyma (482), pancreatic duct (482) \\
\cline{2-3}
 & Pancreas pathology & pancreatic ductal adenocarcinoma lesion (482), pancreas tumor (423) \\
\cline{2-3}
 & Pancreas vessels & pancreatic veins (482), pancreatic arteries (482) \\
\cline{2-3}
 & Spleen & spleen (6823) \\
\cline{2-3}
 & Biliary system & gallbladder (5606), common bile duct (482), biliary system (20) \\
\cline{2-3}
 & Stomach and duodenum & stomach (5840), duodenum (4073) \\
\cline{2-3}
 & Small intestine & jejunum and ileum (3600), small bowel (50) \\
\cline{2-3}
 & Large intestine & colon (3506) \\
\cline{2-3}
 & Colon pathology & colon cancer primaries (126) \\
\cline{2-3}
 & Kidneys & left kidney (5666), right kidney (5603), kidney (3972) \\
\cline{2-3}
 & Kidney pathology & left kidney cyst (1228), right kidney cyst (1228), kidney tumor (559), kidney cyst (489), kidney masses (489) \\
\cline{2-3}
 & Adrenal glands & left adrenal gland (5423), right adrenal gland (5423), adrenal glands (140), left adrenal tumor (20), right adrenal tumor (20) \\
\cline{2-3}
 & Aorta abdominal & aorta (5057), descending aorta (135), aortic arch (135), abdominal aorta (120), abdominal aorta segment 1 (120), ascending aorta (segment) (120), aortic arch false lumen (120), abdominal aorta false lumen (120), abdominal aorta segment 1 false lumen (120), ascending aorta false lumen (120), aortic intimal tear (120), aorta vessel tree (56), thoracic aorta (30) \\
\cline{2-3}
 & Celiac axis & celiac trunk (533), splenic artery (121), celiac trunk / artery of the gut (120), celiac artery (50), coeliac trunk (1), splenic vein (1), left gastric artery (1), right gastric artery (1), short gastric veins (1), right gastroepiploic vein (1) \\
\cline{2-3}
 & Hepatic vessels & hepatic vessels (494), hepatic artery (120), common hepatic artery (1), gastroduodenal artery (1), hepatic artery proper (1), right hepatic artery (1), left hepatic artery (1), separate segment 5/6 vein (variant) (1), right hepatic vein (segment 7 only here, variant) (1), middle hepatic vein (1), left hepatic vein (1) \\
\cline{2-3}
 & Portal venous system & portal vein and splenic vein (2947), portal splenic vein (1597), portal veins (191), portal vein (21), venous system (20), right portal vein (1), left portal vein (1), jejunal branches of the superior mesenteric vein (1), ileal branches of the superior mesenteric vein (1), superior posterior pancreaticoduodenal vein (1), right colic vein (1), middle colic vein (1), pancreaticoduodenal veins (plexus) (1) \\
\cline{2-3}
 & Mesenteric vessels & superior mesenteric artery (171), inferior mesenteric artery (2), inferior mesenteric vein (2), superior mesenteric vein (1) \\
\cline{2-3}
 & Renal vessels & left renal artery (171), right renal artery (171), renal artery (70), renal vein (70), left renal vein (1), right renal vein (1) \\
\cline{2-3}
 & Inferior vena cava & inferior vena cava (5016) \\
\cline{2-3}
 & Gonadal vessels & left gonadal vein (2), right gonadal vein (1), a double/fenestrated right gonadal (1) \\
\cline{2-3}
 & Other abdominal vessels & left ascending lumbar vein (1), hemiazygos vein (1) \\
\cline{2-3}
 & Abdominal wall & abdominal tissue (533), fat (533) \\
\cline{2-3}
 & Other abdominal & stones (20), artery (20), metal (20), metastasectomy site (20) \\
\midrule
Pelvis & Bladder & urinary bladder (5098), bladder (126) \\
\cline{2-3}
 & Prostate & prostate (3003), prostate peripheral zone (269), prostate central gland (237), prostate transitional zone (32) \\
\cline{2-3}
 & Female reproductive & cervix (949), uterus (583), prostate and uterus (360) \\
\cline{2-3}
 & Gonads & gonads (886) \\
\cline{2-3}
 & Rectum & rectum (1281) \\
\cline{2-3}
 & Pelvic bones & sacrum (5720), left hip bone (5720), right hip bone (5720), left femoral head (604), right femoral head (604) \\
\cline{2-3}
 & Iliac vessels & left iliac artery (4514), right iliac artery (4514), left iliac vena (4514), right iliac vena (4514), right common iliac artery (172), left common iliac artery (172), right internal iliac artery (51), left internal iliac artery (51), right external iliac artery (51), left external iliac artery (51), right common iliac vein (2), left common iliac vein (1), right external iliac vein (1), left external iliac vein (1), right internal iliac vein (1), left internal iliac vein (1), right inferior epigastric artery (1), left inferior epigastric artery (1), right epigastric vein (1), left epigastric vein (1), right superficial epigastric vein(s) (1), left superficial epigastric vein(s) (1), middle sacral artery (1), middle sacral vein (1), right deep circumflex iliac artery (1), left deep circumflex iliac artery (1) \\
\cline{2-3}
 & Gluteal vessels & right superior gluteal vein (1), left superior gluteal vein (1), right inferior gluteal vein (1), left inferior gluteal vein (1), right inferior gluteal artery (1), left inferior gluteal artery (1), left superior gluteal artery (1) \\
\cline{2-3}
 & Femoral vessels & right femoral artery (121), left femoral artery (121), right common femoral artery (1), left common femoral artery (1), right common femoral vein (1), left common femoral vein (1), right femoral vein (1), left femoral vein (1), right deep femoral vein (1), right deep femoral artery (1), left deep femoral artery (1), right great saphenous vein (1), left great saphenous vein (1), left lateral circumflex vein (1), left lateral circumflex artery (1), right lateral circumflex artery (1), descending branch of the right lateral circumflex artery (1) \\
\cline{2-3}
 & Pelvic visceral vessels & right inferior vesical veins (1), left inferior vesical veins (1), right obturator veins (1), left obturator veins (1), right internal pudendal artery and vein (1), left internal pudendal artery and vein (1), right obturator artery (1), left obturator artery (1), right pubic veins (1), left ureter (1), right ureter (1), vesical plexus vein branch (only left side shown) (1) \\
\cline{2-3}
 & Rectal vessels & superior rectal artery (1), left colic artery (1), superior sigmoid artery (part of the sigmoid branches) (1), superior rectal vein (with branches) (1), sigmoid veins (1), left colic vein (1), left sacral veins (1), inferior rectal veins (1), right inferior rectal veins (1), left inferior rectal vein (1) \\
\cline{2-3}
 & Pelvic pathology & fdg-avid tumor (1159), psma-avid tumor (597), prostate tumor (188), clinically insignificant prostate lesion (140), clinically significant prostate lesion (140) \\
\midrule
Spine & Cervical vertebrae & c1 vertebra (4504), c2 vertebra (4504), c3 vertebra (4504), c4 vertebra (4504), c5 vertebra (4504), c6 vertebra (4504), c7 vertebra (4504) \\
\cline{2-3}
 & Thoracic vertebrae & t9 vertebra (5546), t10 vertebra (5546), t11 vertebra (5546), t12 vertebra (5546), t7 vertebra (5099), t8 vertebra (5099), t1 vertebra (4559), t2 vertebra (4559), t3 vertebra (4559), t4 vertebra (4559), t5 vertebra (4559), t6 vertebra (4559), t13 vertebra (80) \\
\cline{2-3}
 & Lumbar vertebrae & l1 vertebra (5546), l2 vertebra (5546), l3 vertebra (5546), l4 vertebra (5546), l5 vertebra (5546), l6 vertebra (2743) \\
\cline{2-3}
 & Sacral & s1 vertebra (1228) \\
\cline{2-3}
 & Spinal cord and canal & spinal cord (5433), spinal canal (1383) \\
\cline{2-3}
 & Intervertebral discs & intervertebral discs (616), l5 s1 disc (447), l4 l5 disc (447), l3 l4 disc (447), l2 l3 disc (447), l1 l2 disc (447), t12 l1 disc (447), t11 t12 disc (447), t10 t11 disc (447), t9 t10 disc (447), intervertebra disc (16) \\
\cline{2-3}
 & Other spine & vertebrae (616) \\
\midrule
Extremities & Upper arm & left humerus (3974), right humerus (3974) \\
\cline{2-3}
 & Forearm & left radius (1597), right radius (1597), left ulna (1597), right ulna (1597) \\
\cline{2-3}
 & Hand & carpal left (1597), carpal right (1597), left metacarpal (1597), right metacarpal (1597), left fingers (1597), right fingers (1597) \\
\cline{2-3}
 & Femur & left femur (3974), right femur (3974), femur (100), left femur bone (78), right femur bone (78) \\
\cline{2-3}
 & Patella & left patella (1597), right patella (1597), left patella bone (78), right patella bone (78) \\
\cline{2-3}
 & Leg bones & left tibia (1597), right tibia (1597), left fibula (1597), right fibula (1597), tibia (100), left tibia bone (78), right tibia bone (78), left fibula bone (78), right fibula bone (78) \\
\cline{2-3}
 & Ankle bones & left tarsal (1597), right tarsal (1597), left calcaneous bone (78), right calcaneous bone (78), left talus bone (78), right talus bone (78) \\
\cline{2-3}
 & Foot bones & left metatarsal (1597), right metatarsal (1597), left toes (1597), right toes (1597), left cuboid bone (78), right cuboid bone (78), left intermediate cuneiform bone (78), right intermediate cuneiform bone (78), left lateral cuneiform bone (78), right lateral cuneiform bone (78), left medial cuneiform bone (78), right medial cuneiform bone (78), left navicular bone (78), right navicular bone (78), left phalanges bone (78), right phalanges bone (78) \\
\cline{2-3}
 & Pelvic bones & left pelvis bone (78), right pelvis bone (78) \\
\cline{2-3}
 & Femoral cartilage & femoral articular cartilage (100), left femur distal cartilage (78), right femur distal cartilage (78), left femur head cartilage (78), right femur head cartilage (78) \\
\cline{2-3}
 & Patellar cartilage & left patella cartilage (78), right patella cartilage (78) \\
\cline{2-3}
 & Tibial cartilage & tibial articular cartilage (100), left tibia distal cartilage (78), right tibia distal cartilage (78), left tibia lateral cartilage (78), right tibia lateral cartilage (78), left tibia medial cartilage (78), right tibia medial cartilage (78) \\
\cline{2-3}
 & Other cartilage & left pelvis acetabulum cartilage (78), right pelvis acetabulum cartilage (78), left talus cartilage (78), right talus cartilage (78) \\
\cline{2-3}
 & Knee ligaments & left acl ligament (78), right acl ligament (78), left lcl ligament (78), right lcl ligament (78), left mcl ligament (78), right mcl ligament (78), left pcl ligament (78), right pcl ligament (78) \\
\midrule
Musculo-skeletal & Paraspinal muscles & left autochthon (4514), right autochthon (4514), left longissimus thoracis (97), right longissimus thoracis (97), right spinalis thoracis (1), left spinalis thoracis (1), right ilicostalis lumborum (1), left iliocostalis lumborum (1) \\
\cline{2-3}
 & Abdominal muscles & right quadratus lumborum (1), left quadratus lumborum (1), right rectus abdominis (1), left rectus abdominis (1), right transversus abdominis (1), left transversus abdominis (1), right external oblique (1), left external oblique (1), right internal oblique (1), left internal oblique (1) \\
\cline{2-3}
 & Gluteal muscles & left gluteus maximus (4611), right gluteus maximus (4611), left gluteus medius (4611), right gluteus medius (4611), left gluteus minimus (4071), right gluteus minimus (4071), left gluteus maximus muscle (78), right gluteus maximus muscle (78), left gluteus medius muscle (78), right gluteus medius muscle (78), left gluteus minimus muscle (78), right gluteus minimus muscle (78) \\
\cline{2-3}
 & Hip flexors & left iliopsoas (4610), right iliopsoas (4610), right psoas major muscle (78), left psoas major muscle (78), right illiacus muscle (78), left illiacus muscle (78), right psoas (1), left psoas (1), right iliacus (1), left iliacus (1) \\
\cline{2-3}
 & Hip rotators & right piriformis muscle (78), left piriformis muscle (78), right superior gemellus muscle (78), left superior gemellus muscle (78), right inferior gemellus muscle (78), left inferior gemellus muscle (78), right obturator internus muscle (78), left obturator internus muscle (78), right obturator externus muscle (78), left obturator externus muscle (78), left quadratis femoris muscle (78), right quadratus femoris muscle (78), right piriformis (1), left piriformis (1), right superior gemellus (1), left gemellus superior (1), right inferior gemellus (1), left gemellus inferior (1), right obturator internus (1), left obturator internus (1), right obturator externus (1), left obturator externus (1), right quadratus femoris (1) \\
\cline{2-3}
 & Hip abductors & right tensor fasciae latae muscle (78), left tensor fasciae latae muscle (78), right tensor fascia lata (1), left tensor fascia lata (1) \\
\cline{2-3}
 & Pelvic floor & right ischiocavernosus (1), left ischiocavernosus (1) \\
\cline{2-3}
 & Quadriceps & right rectus femoris muscle (78), left rectus femoris muscle (78), right vastus intermedius muscle (78), left vastus intermedius muscle (78), right vastus lateralis muscle (78), left vastus lateralis muscle (78), left vastus medialis muscle (78), right vastus medialis muscle (78), right rectus femoris (1), left rectus femoris (1), right vastus intermedius (1), left vastus intermedius (1), right vastus lateralis (1), left vastus lateralis (1) \\
\cline{2-3}
 & Other anterior thigh & right sartorius muscle (78), left sartorius muscle (78), right sartorius (1), left sartorius (1) \\
\cline{2-3}
 & Adductors & right pectineus muscle (78), left pectineus muscle (78), right gracilis muscle (78), left gracilis muscle (78), right adductor longus muscle (78), left adductor longus muscle (78), right adductor brevis muscle (78), left adductor brevis muscle (78), right adductor magnus muscle (78), left adductor magnus muscle (78), right pectineus (1), left pectineus (1), right gracilis (1), left gracilis (1), right adductor longus (1), left adductor longus (1), right adductor brevis (1), left adductor brevis (1), right adductor minimus (1), left adductor minimus (1), right adductor magnus (1), left adductor magnus (1) \\
\cline{2-3}
 & Hamstrings & right biceps femoris long head muscle (78), left biceps femoris long head muscle (78), right biceps femoris short head muscle (78), left biceps femoris short head muscle (78), right semitendinosus muscle (78), left semitendinosus muscle (78), right semimembranosus muscle (78), left semimembranosus muscle (78), right biceps femoris (long head) (1), left biceps femoris (long head) (1), right semitendinosus (1), left semitendinosus (1), right semimembranosus (1), left semimembranosus (1) \\
\cline{2-3}
 & Calf posterior & left gastrocnemius lateral muscle (78), right gastrocnemius lateral muscle (78), left gastrocnemius medial muscle (78), right gastrocnemius medial muscle (78), left soleus muscle (78), right soleus muscle (78), left plantaris muscle (78), right plantaris muscle (78), left popliteus muscle (78), right popliteus muscle (78) \\
\cline{2-3}
 & Calf anterior & left tibialis anterior muscle (78), right tibialis anterior muscle (78), left extensor digitorum longus muscle (78), right extensor digitorum longus muscle (78), left extensor hallucis longus muscle (78), right extensor hallucis longus muscle (78) \\
\cline{2-3}
 & Calf deep posterior & left tibialis posterior muscle (78), right tibialis posterior muscle (78), left flexor digitorum longus muscle (78), right flexor digitorum longus muscle (78), left flexor hallucis longus muscle (78), right flexor hallucis longus muscle (78) \\
\cline{2-3}
 & Calf lateral & left peroneus longus muscle (78), right peroneus longus muscle (78) \\
\cline{2-3}
 & General musculoskeletal & bones (2070), skin (886), muscles (533) \\
\midrule
Vascular & Aorta and branches & aortic arch (135), descending aorta (135), abdominal aorta (120), innominate artery (120), innominate artery segment 1 (120), innominate artery segment 1 false lumen (120), right common iliac artery false lumen (120), left common iliac artery false lumen (120), left common carotid artery false lumen (120), left subclavian artery false lumen (120), descending aorta false lumen (120), ascending aorta (101), aorta zone 0 (50), aorta zone 1 (50), aorta zone 2 (50), aorta zone 3 (50), aorta zone 4 (50), aorta zone 5 (50), aorta zone 6 (50), aorta zone 7 (50), aorta zone 8 (50), aorta zone 9 (50), thoracic aorta (30) \\
\cline{2-3}
 & Venous system & inferior vena cava (5016), superior vena cava (1289), portal vein (21), azygos vein (2), splenic vein (1) \\
\cline{2-3}
 & Vascular pathology & true lumen (100), false lumen (100), false lumen thrombus (100), false aortic lumen (39), true aortic lumen (39), pulmonary embolism (35) \\
\midrule
General pathology & Tumors and lesions unspecified & lesion (1093), malignant melanoma tumor (670), fdg-avid soft tissue sarcoma (42), tumor (unspecified) (20) \\
\end{longtable}
\twocolumn

\begin{table}[htbp]
\centering
\begin{tabular}{lrr}
\hline
\textbf{Region} & \textbf{Masks} & \textbf{Labels} \\
\hline
Chest             & 181,095 & 74  \\
Brain             & 155,141 & 280 \\
Head and Neck     & 131,434 & 177 \\
Spine             & 135,132 & 41  \\
Abdomen           & 106,661 & 118 \\
Extremities       & 55,330  & 80  \\
Musculoskeletal   & 54,504  & 150 \\
Pelvis            & 52,323  & 93  \\
Heart             & 27,950  & 36  \\
Vascular          & 8,723   & 34  \\
General Pathology & 1,825   & 4   \\
\hline
\textbf{Total}    & \textbf{910,118} & \textbf{1,087} \\
\hline
\end{tabular}
\label{Tab}
\caption{\textbf{Summary of unique label masks by anatomical region.} 
Aggregate counts of label masks and corresponding labels across all 158 semantic datasets used to train VoxTell. 
Each dataset defines its own label dictionary; the number of label masks is computed as the number of labels multiplied by the number of scans, regardless of whether a structure is present in every scan.}
\end{table}

% \begin{figure}[t]
%     \centering
%     \begin{tikzpicture}[scale=1]
%         % Calculate angles (starting from top, going clockwise)
%         % MRI: 54.93% = 197.748 degrees
%         % CT: 36.69% = 132.084 degrees  
%         % PET: 8.37% = 30.132 degrees
        
%         % Draw pie slices (counterclockwise from 90°)
%         \fill[mri] (0,0) -- (90:2.5) arc (90:-107.748:2.5) -- cycle;
%         \fill[ct] (0,0) -- (-107.748:2.5) arc (-107.748:-239.832:2.5) -- cycle;
%         \fill[pet] (0,0) -- (-239.832:2.5) arc (-239.832:-270:2.5) -- cycle;
        
%         % Add numbers inside slices
%         \node[black, font=\small] at (-8.874:1.5) {25{,}913};
%         \node[black, font=\small] at (-173.79:1.5) {17{,}307};
%         \node[black, font=\small] at (-254.898:1.5) {3{,}949};
        
%         % Draw legend
%         \fill[mri] (3.5,1.5) rectangle (4,1.7);
%         \node[anchor=west] at (4.1,1.6) {\MRI{} (54.93\%)};
        
%         \fill[ct] (3.5,0.75) rectangle (4,0.95);
%         \node[anchor=west] at (4.1,0.85) {\CT{} (36.69\%)};
        
%         \fill[pet] (3.5,0) rectangle (4,0.2);
%         \node[anchor=west] at (4.1,0.1) {\PET{} (8.37\%)};
%     \end{tikzpicture}
%     \caption{Distribution of medical imaging modalities in the semantic train and test datasets. The dataset comprises 47{,}169 images across three imaging modalities. Note, that for MRI we count each sub-modality (T1, T2, etc.) separately.}
%     \label{fig:modality_distribution}
% \end{figure}

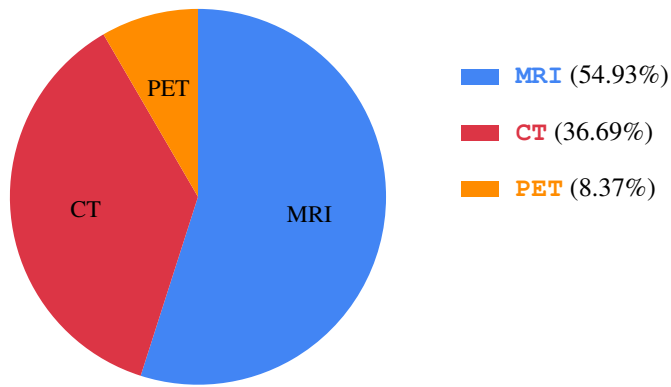
\begin{figure}[t]
    \centering
    \begin{tikzpicture}[scale=1]
        % Calculate angles (starting from top, going clockwise)
        % MRI: 54.93% = 197.748 degrees
        % CT: 36.69% = 132.084 degrees  
        % PET: 8.37% = 30.132 degrees
        
        % Draw pie slices (counterclockwise from 90°)
        \fill[mri] (0,0) -- (90:2.5) arc (90:-107.748:2.5) -- cycle;
        \fill[ct] (0,0) -- (-107.748:2.5) arc (-107.748:-239.832:2.5) -- cycle;
        \fill[pet] (0,0) -- (-239.832:2.5) arc (-239.832:-270:2.5) -- cycle;
        
        % Add numbers inside slices
        \node[black, font=\small] at (-8.874:1.5) {MRI};
        \node[black, font=\small] at (-173.79:1.5) {CT};
        \node[black, font=\small] at (-254.898:1.5) {PET};
        
        % Draw legend
        \fill[mri] (3.5,1.5) rectangle (4,1.7);
        \node[anchor=west] at (4.1,1.6) {\MRI{} (54.93\%)};
        
        \fill[ct] (3.5,0.75) rectangle (4,0.95);
        \node[anchor=west] at (4.1,0.85) {\CT{} (36.69\%)};
        
        \fill[pet] (3.5,0) rectangle (4,0.2);
        \node[anchor=west] at (4.1,0.1) {\PET{} (8.37\%)};
    \end{tikzpicture}
    \caption{Distribution of medical imaging modalities in the semantic train and test datasets.}
    \label{fig:modality_distribution}
\end{figure}


\end{document}